\pdfoutput=1

\documentclass[10pt,twocolumn,letterpaper]{article}

\usepackage[pagenumbers]{cvpr}

\usepackage{pifont}
\usepackage[table,dvipsnames]{xcolor}    %
\usepackage{xparse}
\usepackage{xstring}                     %
\definecolor{baseline}{gray}{0.9}        %
\definecolor{monoirnn}{rgb}{1.0,0.95,0.80}   %
\definecolor{monoirtexgen}{rgb}{0.80,0.90,1.0}%

\newcommand{\mypara}[1]{\noindent\textbf{#1}}

\usepackage[ruled]{algorithm2e} %

\SetAlFnt{\small}
\SetAlCapFnt{\small}
\SetAlCapNameFnt{\small}
\SetAlCapHSkip{0pt}

\definecolor{cvprblue}{rgb}{0.21,0.49,0.74}
\usepackage[pagebackref,breaklinks,colorlinks,citecolor=cvprblue]{hyperref}

\title{LightGenBench: A Benchmark for 3D Emission Generation}

\author{Dongchen Yang \qquad Xingguang Yan \qquad Manolis Savva\\
Simon Fraser University\\
{\small\url{https://3dlg-hcvc.github.io/lightgenbench/}}
}

\begin{document}

\twocolumn[{%
\renewcommand\twocolumn[1][]{#1}%
\maketitle
\begin{center}
   \centering
   \includegraphics[width=1.0\linewidth]{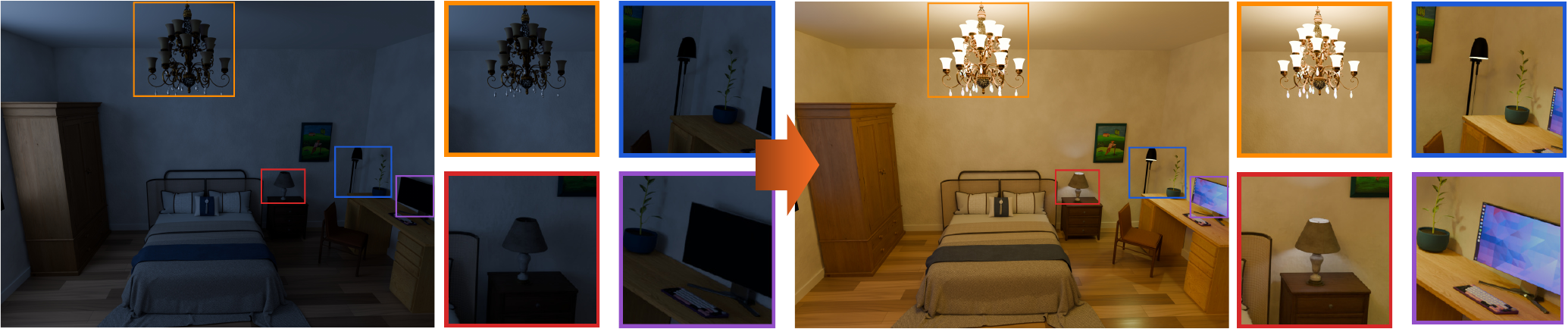}
   \captionof{figure}{\textbf{Emission turns 3D assets into working light sources.} A bedroom scene is rendered with emission off (left) and on (right); the emitters (a chandelier, lamps, and a monitor) are marked by colored boxes and enlarged in the insets. We study how to generate such emission automatically.
   }
   \label{fig:teaser}
\end{center}%
\vspace{0.5em}
}]

\begin{abstract}
Recent generation methods produce 3D assets with detailed geometry and rich textures.
However, these assets are non-emissive: they reflect light but emit none. 
We study a new task, emission generation: predicting spatially varying RGB emission over a 3D asset's surface from its geometry, non-emissive materials, and a rendered view.

To support this task, we introduce LightGenBench. 
Our benchmark makes three contributions. 
First, we curate a dataset of 37k artist-authored emissive assets in UV, voxel, and multi-view form.
Second, we establish baselines across four method families: UV-domain, sparse-voxel, multi-view, and segmentation-based generation.
Third, we design an evaluation protocol that measures where an asset emits and how, complemented by %
user-study comparisons on renderings with varying emission strength.

Across the four families, the sparse-voxel baseline TRELLIS.2-Emission
performs best because generating emission natively in 3D avoids the failure modes of the other representations, such as the fracturing artifacts of UV space and the occluded emitters that multi-view generation cannot see. 
However, TRELLIS.2-Emission only reaches an IoU of 0.271, leaving the task far from solved. 
We hope LightGenBench establishes emission generation as a
benchmarkable problem for 3D generation.
\end{abstract}

\begin{figure*}
  \centering
  \includegraphics[width=\textwidth]{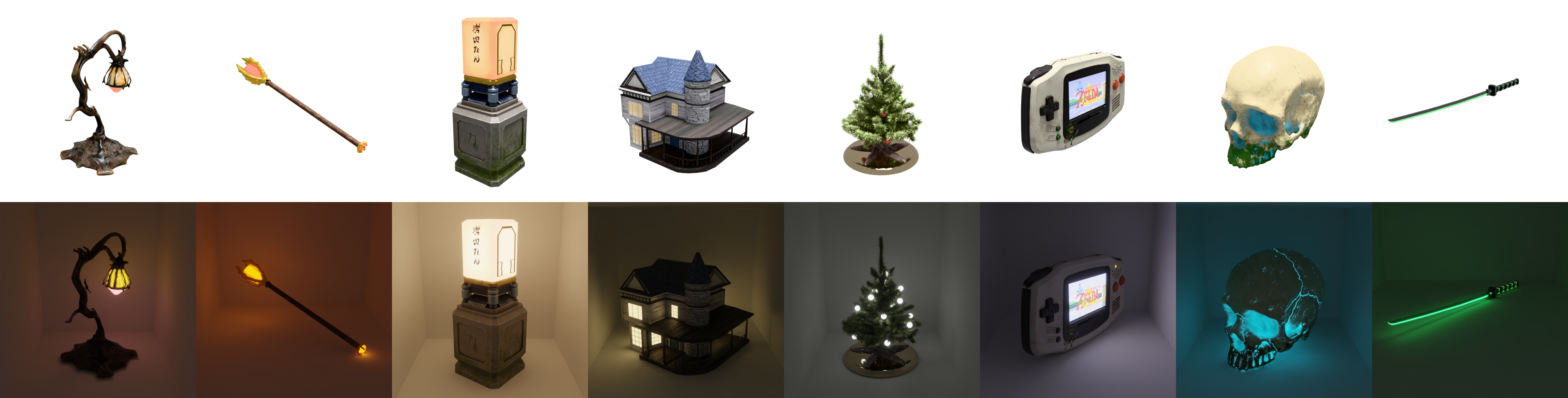}
  \caption{\textbf{Samples from our curated dataset.} Each pair shows an object under standard lighting (top) and the same object rendered in a dark room, lit only
  by its own emission (bottom). Apart from illuminating the object's surroundings (the
  lamps, the lit house), emission also decorates (the glowing cracks, the
  blade, the tree's lights) and informs (the screen's content).
  \looseness=-1
  }
  \label{fig:dataset_gallery}
\end{figure*}

\section{Introduction}

Recent generation methods can produce 3D assets with detailed geometry and rich textures~\cite{xiang2025trellis2,huang2025material,zhao2025hunyuan3d, yan2024object}.
In terms of the rendering equation~\cite{kajiya1986rendering}, existing 3D asset generation methods only generate the reflectance term while leaving the emission term $L_{\text{e}}$ untouched: they can generate a lamp, but the lamp cannot glow.
Emission lets an object light its surroundings rather than only be lit; in \cref{fig:teaser}, the room is dark until its chandelier, lamps, and screen emit.
A common workaround adds a separate point light near the object, as Infinigen Indoors~\cite{infinigen2024indoors} does for its generated lamps, but this degrades realism and the light does not automatically travel with the object across scenes.\looseness=-1

To make the object the light source, we formulate the task of emission generation: given a 3D asset's geometry, its non-emissive materials, and a rendered view, we generate a spatially varying RGB emission for the object.  The emission is parameterized by the standard material model~\cite{gltf2spec}.

Tackling emission generation systematically has three key challenges.
First, no public dataset exists for emission generation, and curating one is non-trivial.
Emissive assets are scarce, and many assets carrying an emissive material are not plausible light sources: artists often copy the albedo into the emission channel to render an asset at full brightness, among other shading workarounds.
While these workarounds result in visually plausible lighting effects, it is hard to recover the actual light sources with simple rules.

Second, there is no established methodology for this task.
The closest are texture generation methods, which produce spatially varying material parameters such as albedo across several representations: UV atlases, sparse voxel grids, and multi-view images.
Methods from each representation could be repurposed for emission.
However, emission is \textit{extremely sparse}, exactly zero over most of a surface.
These methods were built for dense reflectance signals, and whether they work well on such a sparse signal is unclear.
An alternative approach is to try to segment out the sparse emissive regions.  
While recent work has started to investigate emission generation using texture generation~\cite{zhang2026towards}, there is little work tackling this problem and there has been
no systematic comparison of approaches.

Third, evaluation is hard.
In a well-lit rendered image, emitted and reflected light mix throughout the scene.
Emission is easiest to judge through its effect on the environment and through change, as when a lamp switches on.
Hence, having just a well-lit, static rendering is insufficient to test the generated emission.
Instead, it is important to evaluate renderings with different emissive strengths.

We introduce \textbf{LightGenBench}, a benchmark that addresses these difficulties with the following contributions:
\textbf{(i)~a curated dataset} (\cref{fig:dataset_gallery}) of artist-authored emissive 3D assets from TexVerse~\cite{zhang2025texverse}, filtered for genuine light sources by heuristics and a VLM;
\textbf{(ii)~a baseline study} that repurposes representative methods from four families: UV-domain texture generation with TEXGen~\cite{yu2024texgen}, sparse-voxel latent generation with TRELLIS.2~\cite{xiang2025trellis2}, multi-view texture generation as adopted by concurrent work~\cite{zhang2026towards}, and segmentation-based prediction with SegviGen~\cite{segvigen}, where we apply its predicted mask to the input albedo;
\textbf{(iii)~an evaluation protocol} that measures \emph{where} an asset emits (IoU of the emissive regions) and \emph{how} it emits (MAE and PSNR), complemented by user-study comparisons on videos with dynamically varying emission strength, from dim to bright, which better reveal the light source.

Across the four families, the sparse-voxel TRELLIS.2-Emission baseline performs best, indicating that generating emission natively in 3D is currently the strongest design choice.
However, even the best baseline reaches an IoU of only 0.271: it often misplaces the emissive region, leaving the task largely unsolved.
The other representations still have advantages of their own: UV textures are what artists edit, and multi-view generation builds directly on the success of 2D generative models.
Combining native 3D generation with these strengths may be a promising direction.

\section{Related Work}
\label{sec:related_work}
\mypara{Texture generation for 3D objects.}
Texture generation for 3D assets has been studied for a long time, with works exploring various directions such as procedural methods \cite{perlin1985image}, GAN-based generation \cite{siddiqui2022texturify}, CLIP-guided optimization~\cite{jain2022zero, michel2022text2mesh, mohammad2022clip}, and score distillation from 2D diffusion priors~\cite{chen2023fantasia3d, zhang2024dreammat}. Recent diffusion-based methods generate color textures or non-emissive material textures. They can be categorized based on the spatial representations: multi-view projection, UV-space, and 3D-domain methods.

Text2Tex~\cite{chen2023text2tex} and TEXTure~\cite{richardson2023texture} leverage depth-to-image diffusion models to generate multi-view images. The generated images are then projected back onto the mesh texture. 
Paint3D~\cite{zeng2024paint3d} further introduces a refinement stage to remove illumination artifacts and inpaint missing regions in UV space.
To address view inconsistencies caused by sequential inpainting, subsequent works adopt synchronized multi-view generation~\cite{cheng2025mvpaint, yan2025flexpainter, shao2025mvpainter, zhao2025hunyuan3d}. Material Anything~\cite{huang2025material}, MaterialMVP~\cite{zhang2025materialmvp}, and Hunyuan3D-2.1~\cite{hunyuan3d2025} follow a similar pipeline, but generate non-emissive material maps in multi-view images and then project them to 3D.

In contrast to projection-based methods, TEXGen~\cite{yu2024texgen} and  SeqTex~\cite{yuan2025seqtex} synthesize color textures directly in UV space. Object Images~\cite{yan2024object} encodes geometry and non-emissive materials jointly as a multi-chart UV image and generates complete objects with image diffusion.

As for the 3D-domain methods, Texturify~\cite{siddiqui2022texturify} generates textures directly on the mesh surface.
TRELLIS.2~\cite{xiang2025trellis2} stores non-emissive material parameters in voxels and generates them in a 3D latent space. 
SegviGen~\cite{segvigen} repurposes TRELLIS.2 for part
segmentation, predicting per-voxel labels instead of materials.
Although it does not generate appearance, it is relevant to our task:
emission is sparse and its color usually coincides with the albedo, so
generating emission largely reduces to segmenting the emissive regions.

Across all three representations above, the generated materials are reflectance
parameters: none of these methods explicitly generate emissive properties, so the generated assets can only reflect light.
Apart from a preliminary poster, 3D Lighter~\cite{shinya20233d}, the only relevant work, EmissionGen~\cite{zhang2026towards}, formulates emission generation as a texture generation task. 
It builds on the multi-view backbone of Hunyuan3D-2.1~\cite{hunyuan3d2025} and is compared only against multi-view baselines~\cite{wang2025mage,shao2025mvpainter}. 
However, emissive surfaces are often occluded from external viewpoints, such as bulbs inside lampshades or LEDs embedded within devices, which makes them difficult to capture through multi-view rendering. 
Other representations may be better suited for emission generation.
Our benchmark systematically evaluates and compares methods across different representations.

\mypara{Object-centric inverse rendering.}
Inverse rendering aims to recover physical properties, such as materials, geometry, and lighting, from images. 
Recent object-centric inverse rendering works~\cite{chen2024intrinsicanything,jin2023tensoir,gao2024relightable,liang2024gs} focus exclusively on recovering non-emissive material parameters (e.g., albedo, roughness, metallic) along with environmental lighting, treating objects as passive reflectors and leaving emission unmodeled. 
Some scene-level works~\cite{wu2023factorized,lin2025iris,zhu20232} do recover emission for indoor scenes. 
However, scene-level observations capture how the emitter illuminates the surroundings. This additional information helps disentangle the emitted light from reflected light. 
Our setting provides only object-centric observations, which lack this information. 
Therefore, inverse rendering is less reliable for recovering emission in object-centric settings. 
This motivates us to treat object-centric emission as a generation problem: rather than recovering emission from observations, we synthesize plausible emission conditioned on a 3D asset's geometry, non-emissive material, and RGB observation.

\section{Dataset Construction}
\label{sec:dataset}

\begin{figure}
    \centering
       \includegraphics[width=1.0\linewidth]{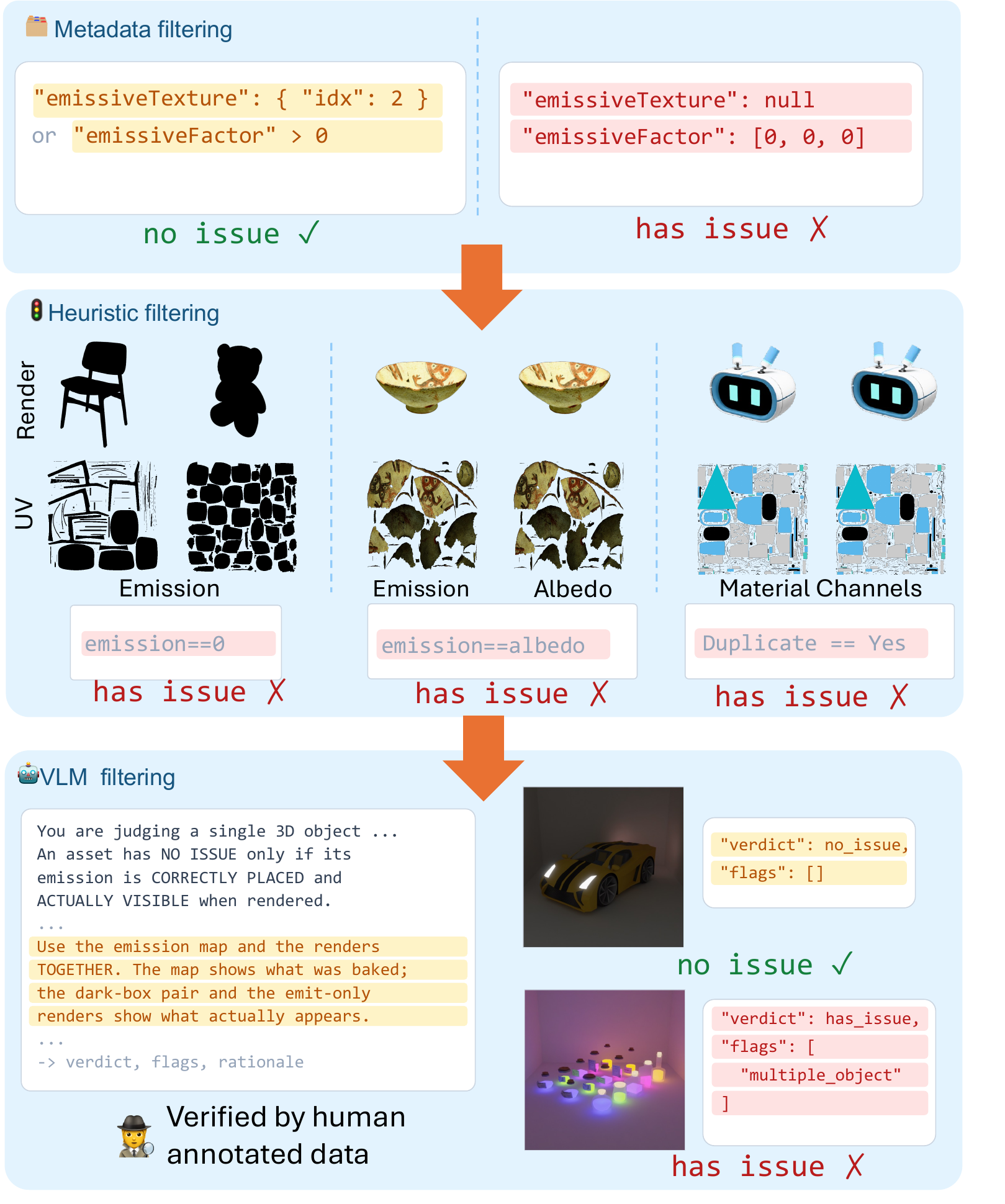}
       \caption{\textbf{Overview of our data filtering pipeline.} 
       (1) Metadata filtering (1st row) keeps assets with an emissive texture or a non-zero emission factor (left) and drops the rest (right). 
       (2) Heuristic filtering (2nd row) removes any asset whose emission map is entirely black (left) or identical to its albedo map (middle), or that duplicates another asset (right).
       (3) VLM filtering (3rd row) rejects non-emissive assets and assets with other defects, e.g., the multi-object asset is rejected. It agrees with human annotations on 87\% of 404 held-out shapes.}
       \label{fig:data_pipeline}
\end{figure}

\begin{figure*}
    \centering
       \includegraphics[width=1.0\linewidth]{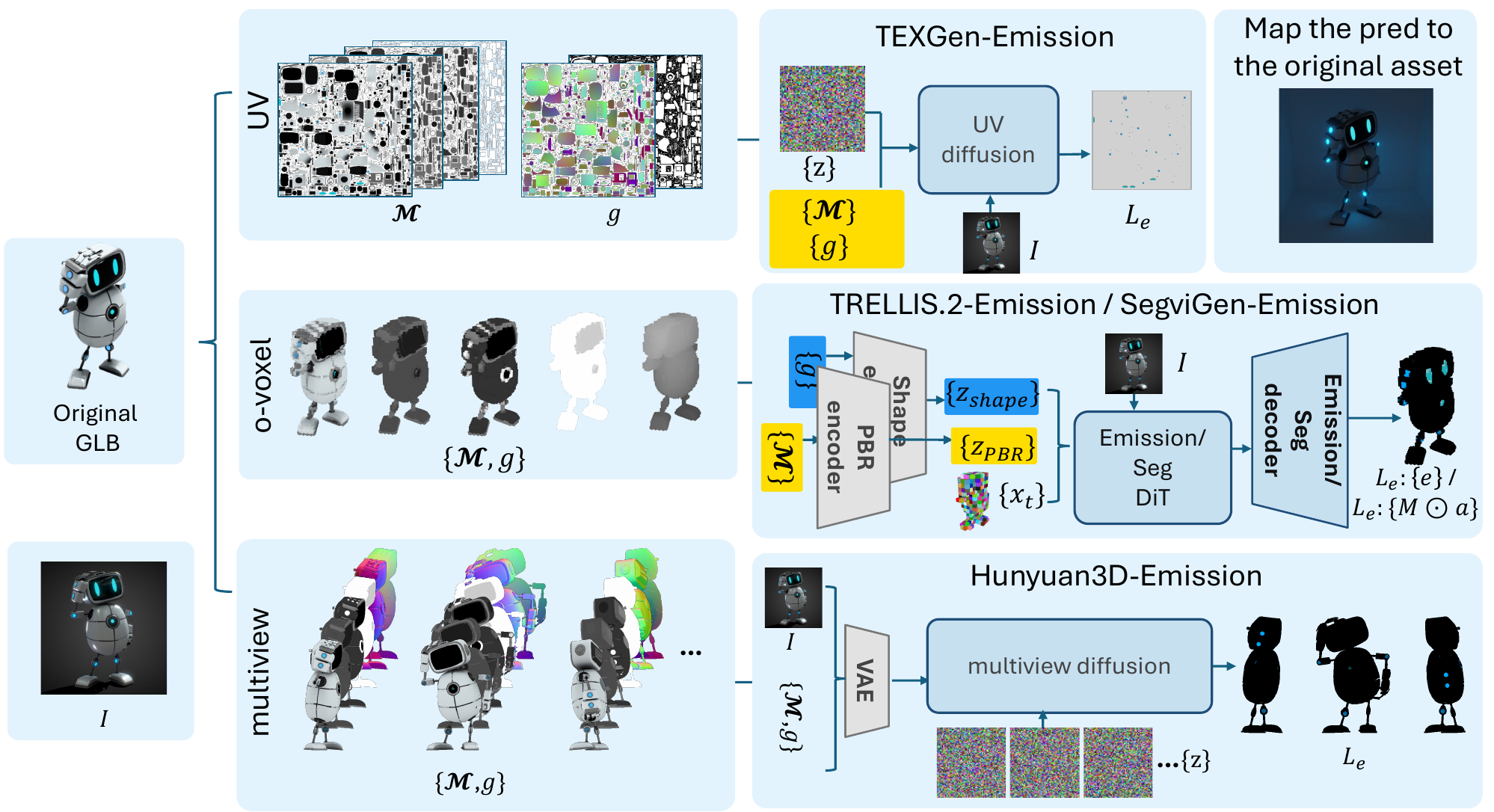}
       \caption{
\textbf{Overview of the four emission-generation methods in our benchmark.} Each method takes the non-emissive materials $\mathcal{M}$, the geometry $g$, and a rendered image $I$, and generates the emission $L_{\text{e}}$ in its own representation: an emission map on the input UV atlas (top), per-voxel emission on the o-voxel grid (middle), or multi-view emission images that are back-projected onto the mesh (bottom). TRELLIS.2-Emission and SegviGen-Emission share the same pipeline and differ in the target: continuous emission ${e}$, or a mask $M$ that is thresholded and multiplied by the albedo ($M\odot a$); their initialization and decoders are detailed in \cref{sec:voxel_model,sec:segvigen}.
       }
       \vspace{-10pt}
       \label{fig:pipeline}
\end{figure*}

Training and evaluating the generative models for emissive materials requires 3D assets with explicit, ground-truth emissive parameters. Though there has been a recent effort~\cite{zhang2026towards} to curate an emissive object dataset from Objaverse~\cite{deitke2023objaverse} as multi-view renderings, this dataset does not enable benchmarking across methods using different representation modalities, and has not been publicly released. We build our own dataset construction workflow (see \cref{fig:data_pipeline}). Our curated dataset is from TexVerse~\cite{zhang2025texverse}, which provides ground-truth emissive materials alongside geometry and non-emissive materials. 
In addition to the multi-view rendering representation in EmissionGen~\cite{zhang2026towards}, we also convert the emissive assets to other representations, including UV-atlas and o-voxels (see \cref{fig:pipeline}).
In all representations, emission is stored in low dynamic range (LDR).
A material's emission is its emissive texture or its emission factor if it has no emissive texture.
We discard the glTF emission strength, since most assets use the default value.

\subsection{Data Filtering Pipeline}
\label{sec:data_filter}
 
We source our 3D assets from TexVerse~\cite{zhang2025texverse}, a collection of 858k 3D models from Sketchfab~\cite{sketchfab}.
Only 158k of them are labeled as having physically based rendering (PBR) materials, but 530k assets contain albedo, metallic, and roughness as a texture or a number, so we source from the entire 858k assets.
TexVerse provides textures at a resolution of at least 1024 pixels, whereas Objaverse~\cite{deitke2023objaverse} provides textures at most 1024 pixels, and about half of Objaverse objects have lower resolutions or no textures.
Emissive regions are small and sparse (\cref{tab:dataset_stats}), so resolution matters for representing them faithfully.
We extract a clean set of emitters in three stages, illustrated in \cref{fig:data_pipeline}: metadata filtering, heuristic filtering, and VLM filtering.

\mypara{Metadata filtering.}
We first detect candidate emitters from each object's metadata: we read the GLB file and keep the assets that have a used material with a non-zero \textit{emission factor} or an emissive texture.
We further require at least one used material to define albedo, metallic, and roughness, as a texture or as a number, since the model takes these as input.
83k of the 858k assets pass the metadata filtering.

\mypara{Heuristic filtering.}
From the candidates after the metadata filtering, we discard the assets that emit nothing, the assets that simply copy the albedo into the emission channel, and duplicate copies of the same asset.
A material that is used by the scene is emissive when its emissive texture is not all black, or its emission factor is non-zero if it has no emissive texture.
We discard an asset when none of its materials emits.
If an asset's emission comes only from surfaces that are fully transparent, we also discard it.
After this non-zero emission check, 78k shapes are left.
Since artists may copy the content from the albedo channel to the emission channel, we further compare the two channels.
The emission of a material is a copy of the albedo when: 1) its emission slot and its albedo slot point to the same image in the file, 2) the two slots point to different images that hold the same picture, or 3) for a material without an albedo texture image, the emission color equals the albedo color.
We discard an asset when the emission of every material it uses is such a copy.
Of the discarded assets, 159 have a human label from our annotation, and 95.6\% of them are labeled as not emissive.
After this step, 73k shapes are left. 
TexVerse also contains re-uploads of the same asset.
We treat two assets as duplicates when they have the same geometry, the same albedo, and the same emission.
We keep one asset from the duplicated asset groups.
After this step, 70k shapes are left.

\mypara{VLM filtering.}
We then render the remaining objects in several formats: a dark-room rendering,
where the object is placed in a room with dim ambient light and rendered with its emission turned
on and off; a multi-view rendering; and a multi-view emit-only rendering, where
only the emissive parts are shown. 
We keep 66k of the 70k shapes for which all renderings are available and the baked emission atlas is not all zero.
Together with the baked emission atlas, these renderings form the input to GPT-5.6 Luna. 
To validate the VLM, we compare it against human annotation
on 802 sampled shapes.
We use 398 of them to refine the prompt: we inspect the shapes on which the VLM
disagrees with the human labels and revise the judging criteria in the
prompt.
It reaches an 87\% alignment rate with the human annotators for the remaining 404 shapes.
We then apply this workflow to the $\sim$66k shapes, resulting in $\sim$37k shapes.

\begin{table}
\centering
\caption{\textbf{Dataset statistics.}
The statistics of our curated dataset.
Per-shape \emph{emissive ratio} denotes the percentage of UV-occupied texels with non-zero emission.}
\begin{tabular}{@{} l r @{}}
\toprule
Statistic & Value \\
\midrule
\# Shapes                              & 36{,}826 \\
\# Categories                          & 105 \\
Train / Val / Test                     & 36{,}426 / 200 / 200 \\
Mean per-shape emissive ratio            & 8.1\% \\
Median per-shape emissive ratio           & 1.4\%  \\
\bottomrule
\end{tabular}%
\label{tab:dataset_stats}
\end{table}

\mypara{Data statistics.} The final dataset contains 36{,}826 emissive shapes spanning 105 categories.
\Cref{tab:dataset_stats} summarizes the statistics of the dataset and \cref{fig:dataset_gallery}
shows examples.
Emission in this dataset is sparse: the emissive area covers 8.1\% of a shape's
surface on average and 1.4\% at the median.
We split the shapes into 36{,}426 for training, 200 for validation, and 200 for
testing. The validation split is used for model selection and the test set
for the reported numbers.

\section{Benchmark Setup}
\label{sec:method}
For our benchmark, we repurpose and evaluate four methods for emission generation (\cref{fig:pipeline}): TEXGen~\cite{yu2024texgen}, TRELLIS.2~\cite{xiang2025trellis2}, and Hunyuan3D-2.1~\cite{hunyuan3d2025} generate emission directly in the UV, voxel, and multi-view domains, respectively, while SegviGen~\cite{segvigen} shares the voxel representation but poses the task as emissive-region segmentation. We include a segmentation method because a key challenge is to localize the emissive regions. The segmentation can be converted to emission by heuristics such as applying the albedo color to the emissive segmentation.

\subsection{Problem Formulation}
\label{sec:problem}

The rendering equation~\cite{kajiya1986rendering} decomposes the outgoing radiance $L_{\text{o}}$ into emitted and reflected components:
\begin{equation*}
\begin{split}
L_{\text{o}}(\mathbf{x}, \omega_{\text{o}}, \lambda) =& {}  L_{\text{e}}(\mathbf{x}, \omega_{\text{o}}, \lambda) \\
 + & \int_{\Omega} f_{\text{r}}(\mathbf{x}, \omega_{\text{i}}, \omega_{\text{o}}, \lambda) L_{\text{i}}(\mathbf{x}, \omega_{\text{i}}, \lambda) (\omega_{\text{i}} \cdot \mathbf{n}) \operatorname{d}\omega_{\text{i}},
\end{split}
\end{equation*}
where $L_{\text{e}}$ is the emitted radiance and the integral term is the reflected radiance $L_{\text{r}}$. 
Prior work in 3D generation has extensively explored the synthesis of non-emissive properties in $L_{\text{r}}$ as discussed in \cref{sec:related_work}, while the emission term $L_{\text{e}}$ has received comparatively little attention. 
We address this gap by focusing on the generation of the emission term: given the geometry $g$, the non-emissive material properties $\mathcal{M}=\{a, m, r,\alpha\}$ (albedo, metallic, roughness, transparency), and a rendered image $I$ of the asset, our goal is to generate the spatially-varying emission $L_{\text{e}}$ on the surface.
$I$ is a single-view image in which emissive regions are only partially visible and often indistinguishable from albedo.

\subsection{UV Domain: TEXGen-Emission}
\label{sec:uv_model}

TEXGen~\cite{yu2024texgen} leverages  progress in 2D image generation ~\cite{podell2023sdxl} and generates textures as images over the UV atlas of a mesh. 
Its backbone model, PointUVNet, interleaves UV-space convolution and attention with sparse 3D point-cloud blocks. 
Given a partial albedo map, an image $I$, and the geometry $g$ of an object, which includes a position map ${P}$ and an occupancy mask ${M}$, it
completes the partial albedo map.
Our TEXGen-Emission (\cref{fig:pipeline}, top middle) follows the same design choice and repurposes the TEXGen model for emission generation in the UV domain.

\mypara{Inputs and outputs.} 
To repurpose the original TEXGen for our emission generation task, we replace the input partial albedo map with complete maps of non-emissive material $\mathcal{M}$ channels, including albedo $a$, metallic $m$, roughness $r$, and transparency $\alpha$. The complete input is
$\{{z}_t \;\Vert\; \mathcal{M}\Vert\; g\}$,
where ${z}_t$ is the noised emission map at timestep $t$, $g$ is the geometry of the object in UV space, which includes the position map ${P}$, and the occupancy mask ${M}$ indicating valid surface pixels. These parameters are concatenated ($\Vert$) along the channel
dimension. 
In addition to the UV-space input, the model is conditioned on a CLIP~\cite{radford2021learning} image embedding of a thumbnail rendering of the asset. 
The model predicts a 3-channel emission map $L_{\text{e}}$ in UV space.

\mypara{Training and inference.}
TEXGen releases inference code only, so we implement training ourselves following the flow-matching formulation of the released model. We supervise the flow velocity with an $\ell_2{+}0.5\,\ell_1$ loss over the occupied UV texels. We work at a UV resolution of $256\times256$ rather than the original $1024\times1024$ to reduce compute, and fix the text prompt ``emission generation'', since our task is not text-conditioned. At inference, we use 50 Euler steps.

\subsection{Voxel Domain: TRELLIS.2-Emission}
\label{sec:voxel_model}

Our voxel-domain method is built on TRELLIS.2~\cite{xiang2025trellis2}, a
two-stage image-to-3D generator over sparse o-voxel grids.
Stage~1 generates
the geometry, and Stage~2 generates the PBR (non-emissive) material. 
Each stage consists of a VAE, which compresses per-voxel attributes
into a latent grid, and a DiT, which generates this latent conditioned on the
input image.
The PBR DiT is additionally conditioned on the shape latent. 
For our TRELLIS.2-Emission (\cref{fig:pipeline}, middle right), we keep these two stages frozen
and add a third emission generation stage. 
We use the pretrained shape VAE and the PBR VAE as
frozen encoders for our conditions. 
Our emission VAE and DiT are initialized using the weights from the pretrained PBR generation modules. We will introduce the emission VAE and DiT below.

\mypara{Emission VAE.}
The encoder for our emission-VAE takes the emission $L_{\text{e}}$ in voxel-space, and the decoder takes $[\text{z}_{\text{e}} \,\Vert\, \text{z}_{\text{PBR}}]$,
and outputs the reconstructed emission. 
$\text{z}_{\text{PBR}}$ is the latent from the frozen PBR encoder, which lets the material $\mathcal{M}$ guide the
reconstruction.
The encoder and decoder are initialized from the pretrained PBR VAE. 
The layers whose channel counts change are initialized from the corresponding
slices of the pretrained weights.

\mypara{Emission DiT.}
The emission DiT generates $\text{z}_{\text{e}}$ with flow matching. 
It is fine-tuned from the pretrained PBR DiT. 
Each latent-voxel token concatenates the noised emission latent $\text{x}_t$, the shape latent $\text{z}_\text{shape}$, and the PBR latent $\text{z}_\text{PBR}$:
$[\,\text{x}_t \,\Vert\, \text{z}_{\text{shape}} \,\Vert\, \text{z}_{\text{PBR}}\,]$.
The shape latent $\text{z}_\text{shape}$ is obtained by encoding the geometry $g$, which is the o-voxel dual grid.
Because of the change in input channels, the input projection is re-initialized. 
The model predicts the velocity of $\text{z}_{\text{e}}$. 
The thumbnail rendering $I$ enters as DINOv3~\cite{simeoni2025dinov3} features as in the original TRELLIS.2.

\mypara{Training and inference.}
\label{sec:trellis2-emission-train-inf}
We train the two modules in sequence. 
The emission VAE is fine-tuned from the pretrained PBR VAE with an $\ell_1$ 
reconstruction loss and a KL term. 
We then fine-tune the emission DiT on the latents this VAE produces with the
standard flow-matching loss~\cite{lipman2022flow}. 
Both stages condition on the PBR latent $\text{z}_{\text{PBR}}$, and the DiT
is additionally conditioned on the shape latent $\text{z}_{\text{shape}}$.
$\text{z}_{\text{shape}}$ and $\text{z}_{\text{PBR}}$ are obtained by encoding the o-voxel dual grid $g$
and the non-emissive material attributes $\{a, m, r, \alpha\}$ with the
frozen pretrained encoders.
At inference, we sample with 50 Euler steps and decode $\text{z}_{\text{e}}$ with
the emission VAE decoder to per-voxel emission on the input occupancy.

\subsection{Multi-view Domain: Hunyuan3D-Emission}

Hunyuan3D-2.1~\cite{hunyuan3d2025} generates PBR textures with multi-view diffusion. 
Given a mesh and a reference image $I$, it denoises the albedo $a$ and metallic--roughness $\{m,r\}$ images of six views. Per-view normal $n$ and position $P$ renderings are concatenated to the noisy latents as geometry $g$ conditions. 
Our Hunyuan3D-Emission (\cref{fig:pipeline}, bottom right) keeps this pipeline but modifies its inputs and outputs.

\mypara{Inputs and outputs.}
Because our goal is to generate emission, we replace the albedo $a$ and metallic--roughness $\{m,r\}$ generation heads with a single emission generation head.
 The per-view renderings of the non-emissive material $\mathcal{M} =\{a,m,r,\alpha\}$, which are albedo $a$, metallic $m$, roughness $r$, and transparency $\alpha$, are encoded by the Stable Diffusion VAE.
These non-emissive material latents are then concatenated with the noised emission latents $z$ and the geometry latent $g$. 
A thumbnail rendering of the asset serves as the reference image. 
It enters as a Stable Diffusion VAE latent and DINOv2 features. 
The model outputs six emission views. 
We back-project them onto the mesh.

\mypara{Training and inference.}
We fine-tune the pretrained Hunyuan3D-2.1 using DDPM with $v$-prediction~\cite{salimans2022progressive}.
At inference, we use 30 Euler-ancestral steps.

\subsection{Segmentation: SegviGen-Emission}
\label{sec:segvigen}
The methods above generate emission as a continuous signal. 
Emission is sparse, and its color often coincides with the albedo. 
For such objects, the challenge is not predicting the emitted color but finding the part that emits. 
We therefore incorporate segmentation for our emission generation task: the target is to generate a mask indicating the emissive region and use the albedo color of this region for emission color.
We start from SegviGen~\cite{segvigen}, which repurposes the TRELLIS.2 texture DiT for part segmentation. 
The original SegviGen generates a per-voxel coloring, and each color corresponds to a segment. 
It shares the architecture and latent space of TRELLIS.2~\cite{xiang2025trellis2}.
Our SegviGen-Emission (\cref{fig:pipeline}, middle right) keeps the SegviGen formulation and reduces the
multi-class coloring to a binary one.

\mypara{Inputs and outputs.}
The model predicts a per-voxel binary mask $M$ over the occupied voxels. 
The emission can be obtained in heuristic ways. Here, we apply the predicted mask to the input albedo, $\hat{L}_{\text{e}} = M \odot a$. 
Following SegviGen, the target is written into the albedo slot of the PBR attribute vector as a binary coloring, $[\,M, M, M, 0, 1, 1\,]$,
and encoded with the pretrained TRELLIS.2 VAE. 
Following SegviGen, the model is conditioned on the shape and PBR latents, and
the thumbnail enters through DINOv3 cross-attention.

\mypara{Training and inference.}
We fine-tune from SegviGen's full-segmentation checkpoint with flow matching as in \cref{sec:trellis2-emission-train-inf}.
At inference, we integrate the velocity field with 12 Euler steps, decode to per-voxel values with the TRELLIS.2 VAE, and threshold to obtain the predicted mask.

\section{Experiments}

We train the baselines of \cref{sec:method} on our curated dataset and compare them under the same protocol.
Since the baselines generate emissions in different representations, we evaluate them in a canonical sampled point cloud.
In addition, we evaluate the baselines with a user study since the ground truth may not be the only plausible emission for the object.

\subsection{Experimental Setup}

All four baselines are trained on the same training split and receive the same input: the geometry, the non-emissive material parameters, and one thumbnail rendering of the asset (its platform preview, rendered with artist-chosen settings). 
During training, we select the best checkpoint on the validation split, and we report all results on the test set.
Since a single prediction for generative baselines is an unstable measurement, we evaluate baselines five times with different random seeds and report the mean across them.

\subsection{Sampled Point Cloud Evaluation}

For each asset, we sample 50{,}000 points uniformly on the input mesh, and the evaluation is conducted on these sampled points.
The reference emission is the emission color of each sampled point on the original asset.
The predicted emission at the sampled point is read from each method's own output by a nearest-neighbor lookup in its own output representation:
the nearest occupied voxel for TRELLIS.2-Emission and SegviGen-Emission, and the nearest atlas texel by 3D position for TEXGen-Emission. 
For Hunyuan3D-Emission, we look up the nearest atlas texel, since its six generated views are back-projected into a UV atlas first. See the supplement for more details.

We evaluate the predictions with intersection over union (IoU) for the emissive mask, and mean absolute error (MAE) and peak signal-to-noise ratio (PSNR) on the emissive color.
IoU measures if the emissive region is generated at the right location. 
We threshold both the prediction and the reference at $1/255$, and report the IoU of the two point sets.
MAE and PSNR measure whether the emitted color is correct. We compute them over all sampled points.
\begin{figure}
    \centering
       \includegraphics[width=1.0\linewidth]{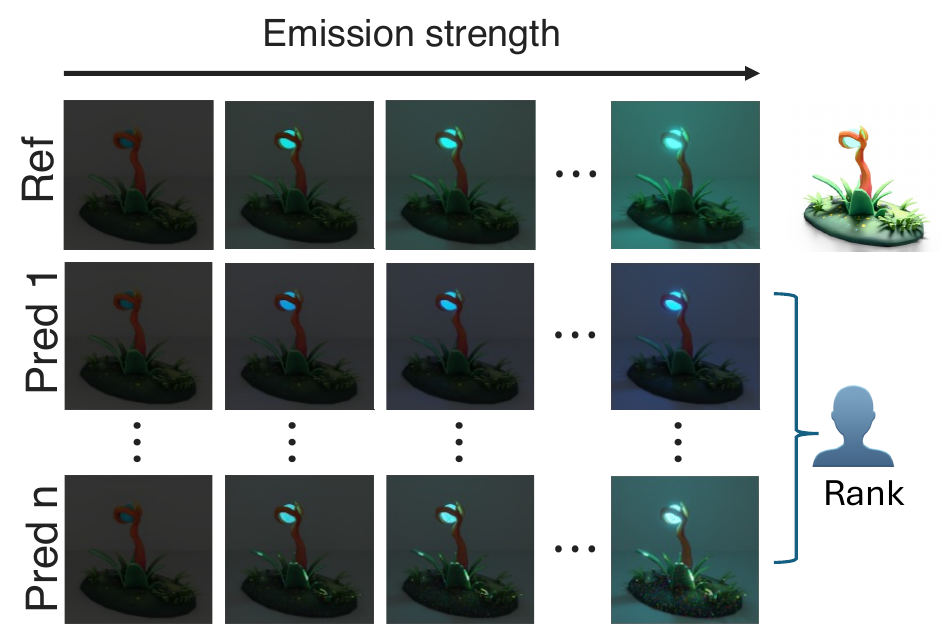}
       \caption{\textbf{User study.} Each asset is rendered in a dark room with each
baseline's predicted emission (Pred 1 to Pred $n$) ramping from 0 to 8. Participants rank the
baselines by plausibility, with the ground-truth emission video and a well-lit
rendering as reference (Ref).}
       \label{fig:user_study}
\end{figure}

\subsection{User Study}

Since generated emission can be plausible without closely matching the ground truth, we additionally conduct a user study.
For each asset, we place it in a dark room and sweep the strength of its predicted emission from 0 to 8, so that the emitter brightens from dim to bright. 
This brightness change makes the things the metrics measure easy to judge by eye:
where the asset emits and what color the light is.
Participants see the ground-truth rendering of the object and the ground-truth video as a
reference, and the videos for the four baselines. 
Participants are asked to rank the baselines by how plausible the emission looks.
The study covers 53 assets and 14 participants.
\Cref{fig:user_study} shows the protocol.

\begin{table}
\centering
\caption{\textbf{Quantitative evaluation on the test set.}
We report the mean over five sampling seeds.}
\label{tab:quant}
\begin{tabular}{@{}lccc@{}}
\toprule
Method & IoU $\uparrow$ & MAE $\downarrow$& PSNR $\uparrow$ \\
\midrule
TEXGen-Emission    & 0.092 & 0.025 & 25.10\\
TRELLIS.2-Emission & \textbf{0.271} & \textbf{0.021} & \textbf{26.81}\\
Hunyuan3D-Emission & 0.117 & 0.022 & 26.46\\
SegviGen-Emission  & 0.147 & 0.041 & 24.50\\

\bottomrule
\end{tabular}
\end{table}

\subsection{Results}
\begin{figure*}
    \centering
       \includegraphics[width=1.0\linewidth]{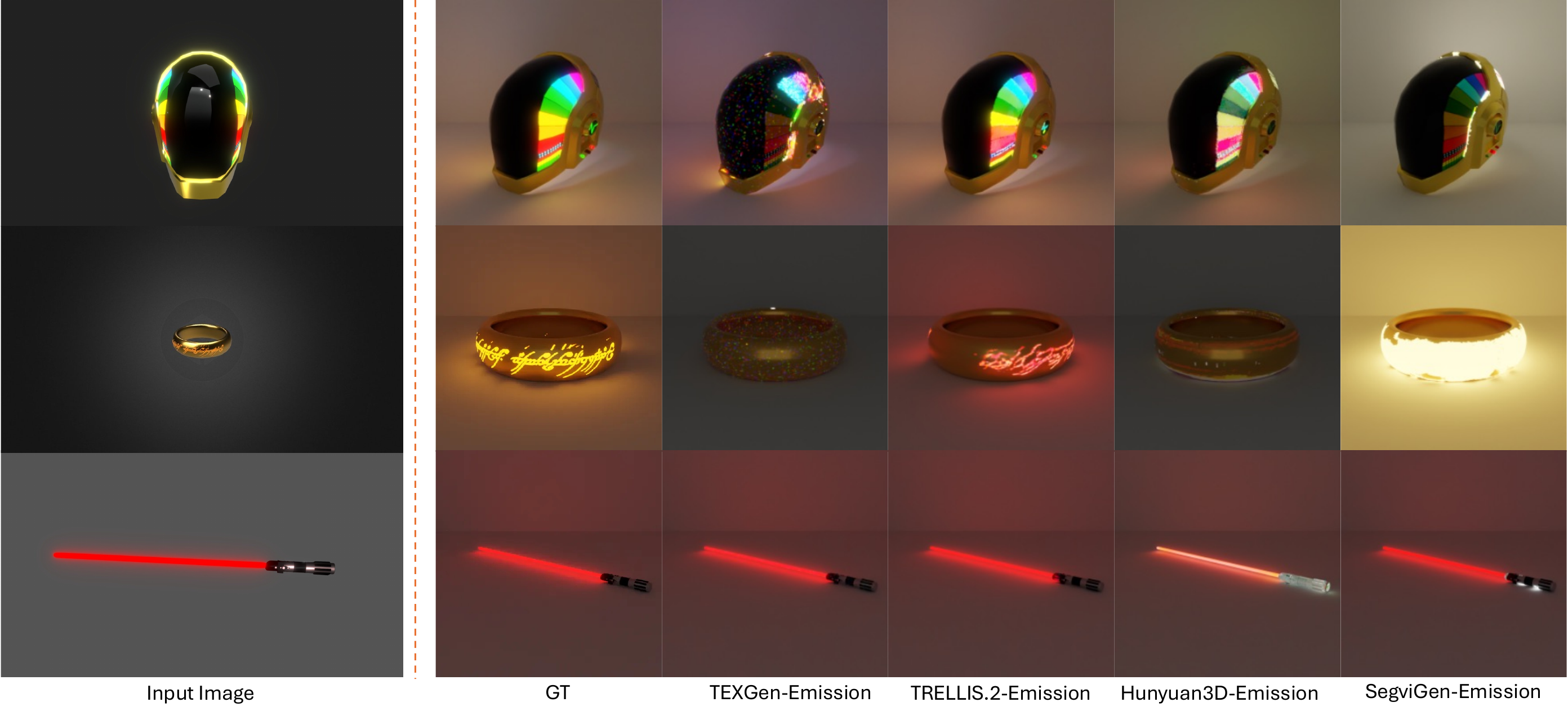}
       \caption{\textbf{Qualitative comparison of the baselines.}
        Each object is rendered in a dark room and lit only by its predicted emission; 
        the leftmost column is the input image, and GT is rendered with ground truth emission.
       \textit{TRELLIS.2-Emission} stays closest to the ground truth on all three assets, but still mislocates emissive surfaces or assigns inaccurate color (the ring’s emissive region is misplaced and it should glow yellow).
       The other baselines show their characteristic failures: 
       \textit{TEXGen-Emission} scatters emission across UV islands (spilled color on the helmet) 
       and fails to keep non-emissive regions at exact zero (speckles on the helmet, faint glowing noise over the ring);
       \textit{Hunyuan3D-Emission} shifts the emission color (the red beam turns white) and lights up regions that should stay dark (the saber handle and the ring band); 
       and \textit{SegviGen-Emission} over-extends or misses emissive regions (the ring's whole band glows, while the helmet's stripes stay unlit).}
       \label{fig:qual_box}
\end{figure*}

We show qualitative results in \cref{fig:qual_box}, where each object is placed in a dark room and lit only by its predicted emission, and quantitative and user study results in \cref{tab:quant,tab:user_study_pref}.
Overall, TRELLIS.2-Emission performs best on all three metrics and in the user study.
However, no method is close to the reference: the best average IoU is only 0.271. 
This indicates that emission generation remains largely unsolved.

\mypara{UV domain generation fractures across UV islands.}
TEXGen-Emission, which predicts emission in UV space, produces fracturing artifacts (the helmet example in \cref{fig:qual_box} 3rd column).
Regions that are close in the UV atlas are generally not close on the surface, but a model operating in UV space may treat them as neighbors.
Emission predicted for one island therefore could spill into the islands around it, whether by overflowing the island boundary or assigning color to a nearby island. Those islands may map back to surface regions that should not emit.

\mypara{Multi-view generation misses occluded emitters.}
Hunyuan3D-Emission generates emission as a set of multi-view renderings. 
If an emissive region is occluded or falls outside the selected views, the correct emission will never be generated. 
For example, the light bulb occluded by the lamp hood can never be observed, and thus the plausible emission can not be generated (see supplement).

\mypara{Sparsity needs to be specially handled.}
Emission is sparse: 
on average, only 8.1\% of a shape's surface is emissive (\cref{tab:dataset_stats}), so most regions in an emission texture are zero.
A model therefore has to produce plausible emission textures with zeros for large regions.
TEXGen-Emission and Hunyuan3D-Emission often generate faint noise on non-emissive surfaces (\cref{fig:qual_box} 3rd and 5th column).
For non-emissive textures, such noise would be barely visible.
For emissive textures, however, every non-zero texel is a light source and glows against the surface around it once rendered, especially in our dark-room renderings.
Therefore, producing exact zeros for non-emissive parts becomes a requirement. Thresholding the output would suppress this noise, but it would also remove genuinely dim emission.
TRELLIS.2-Emission and SegviGen-Emission largely avoid this artifact.
TRELLIS.2-Emission decodes through a VAE trained specifically for emission, and SegviGen-Emission is trained as a segmentation model and outputs a near-binary mask.
TEXGen-Emission, by contrast, denoises the UV map directly and has no mechanism to suppress such noise. 
Hunyuan3D-Emission decodes through the Stable Diffusion VAE, which is trained on natural images where large black regions are rare.

\mypara{Native 3D generation is a promising direction.}
Our two voxel-based baselines, TRELLIS.2-Emission and SegviGen-Emission, obtain the highest IoU in \cref{tab:quant}.
Their remaining errors are mainly misplaced emissive surfaces or wrong colors (\cref{fig:qual_box} 4th and 6th column).
However, the other representations retain advantages of their own: multi-view generation builds directly on the success of 2D diffusion models, and UV textures remain the representation that artists author and edit.
We think that combining native 3D generation with the strengths of the other representations may be a good direction for this task.

\begin{table}
\centering
\caption{\textbf{User study results.} 14 participants each ranked the outputs of all
four baselines on 53 objects. We report the mean rank and the fraction of rankings in which a method
was placed first.}
\label{tab:user_study_pref}
\begin{tabular}{@{}lcc@{}}
\toprule
Method & Mean rank $\downarrow$ & Top-1 rate $\uparrow$ \\
\midrule
TEXGen-Emission     & 3.22 $\pm$ 1.02 & 10.1\%  \\
TRELLIS.2-Emission  & \textbf{1.46 $\pm$ 0.76} & \textbf{67.2\%}  \\
Hunyuan3D-Emission  & 2.50 $\pm$ 0.95 & 14.3\%  \\
SegviGen-Emission   & 2.82 $\pm$ 0.89 & 8.4\%   \\
\bottomrule
\end{tabular}
\end{table}

\section{Conclusion}

We introduced LightGenBench, a benchmark for emission generation.
We provide a curated dataset of emissive assets, baselines from four method families, and an evaluation protocol that measures where an asset emits and how.
Across the four families, the sparse-voxel TRELLIS.2-Emission baseline performs best, suggesting that generating emission natively in 3D is a promising direction, though combining it with the strengths of other representations remains open.

\section*{Acknowledgments}
This work was funded in part by a Canada Research Chair, NSERC Discovery Grants, and enabled by support from the Digital Research Alliance of Canada, and an NVIDIA Academic Grant Award. We thank Denys Iliash, Han-Hung Lee, and Xiaohao Sun for helpful discussions.

\clearpage
{
    \small
    \bibliographystyle{ieeenat_fullname}
    \bibliography{bibliography}
}
\clearpage
\appendix

\definecolor{cvprblue}{rgb}{0.21,0.49,0.74}

\renewcommand{\thesection}{\Alph{section}}

\maketitlesupplementary

\setcounter{section}{0}
\setcounter{figure}{0}
\setcounter{table}{0}
\renewcommand{\thesection}{\Alph{section}}
\renewcommand{\thefigure}{S\arabic{figure}}
\renewcommand{\thetable}{S\arabic{table}}
The supplementary material is organized as follows.
\Cref{sec:supp_dataset} provides more details about the dataset: the definition of each filtering stage (\cref{sec:supp_filtering}), the three representations of the dataset (\cref{sec:supp_representations}), additional statistics (\cref{sec:supp_statistics}),  and a gallery of our curated data (\cref{fig:supp_gallery}).
\Cref{sec:supp_method_training_details} gives the architecture, training, and inference details of TEXGen-Emission (\cref{sec:supp_texgen}), TRELLIS.2-Emission (\cref{sec:supp_trellis2}), Hunyuan3D-Emission (\cref{sec:supp_hunyuan}), and SegviGen-Emission (\cref{sec:supp_segvigen}).
\Cref{sec:supp_evaluation_protocol_details} describes the evaluation protocol: the sampled point cloud evaluation, the dark room rendering, and the user study interface (\cref{fig:user_study_snapshot}).
\Cref{sec:supp_additional_results} reports additional results: emissive-mask precision, recall, and F1 (\cref{tab:supp_prf}), MAE and PSNR within the ground-truth emissive region (\cref{tab:supp_mask_magnitude}), failure cases of the baselines (\cref{fig:multiview_failure,fig:texgen_fracture,fig:supp_hy_noise}), and more qualitative comparisons (\cref{fig:supp_additional_qual}).
\Cref{sec:supp_failre_cases_limitations} discusses limitations and future work. We also provide a supplementary video that introduces the task, the dataset, and the baselines, and shows the qualitative comparisons and failure cases.

\section{Dataset Details}
\label{sec:supp_dataset}
We provide more details on our dataset: the full definition of each filtering stage and more statistics of the final dataset.
\subsection{Data Filtering}
\label{sec:supp_filtering}
\paragraph{Metadata Filtering.}
We read the glTF JSON of all 857k downloaded TexVerse~\cite{zhang2025texverse} assets.
A material declares emission when it has an emissive texture or a non-zero emission factor if it has no emissive texture.
Only materials used by the asset are considered, and 103{,}762 assets (12.1\%) have at least one such material that declares emission.
We further require at least one used material to have albedo, metallic, and roughness defined, since the baselines take these channels as input.
This keeps 83{,}124 assets, which enter the heuristic filtering stage.
Only 36{,}793 of the 103{,}762 emitting assets carry the TexVerse PBR label, so selecting by the label would miss most of them.

\paragraph{Heuristic Filtering.}
The non-zero emission check removes 5{,}404 of the 83{,}124 candidates whose emissive textures are all black and 151 whose emitting surfaces are fully transparent, leaving 77{,}569.
In the albedo copy check, two images hold the same picture when their mean difference is below 1/255 after resizing both to $256^2$; the check removes 4{,}137 assets, leaving 73{,}432.
The duplicate check compares the geometry and the albedo and emission textures, and removes 3{,}490 assets, leaving 69{,}942.

\paragraph{VLM Filtering.}
We first label $\sim$800 shapes by hand in a web annotation tool.
The annotator primarily sees four renderings: (1) the object under normal lighting, (2) an emissive-brightened rendering, (3) the object in a dark box with its emission turned off ($\text{emission strength}=0$), and (4) the object in a dark box with emission on ($\text{emission strength}=4$).
Four more images are also available if the annotator is still not sure based on the primary images: an emit-only rendering, the baked emission map in UV space, and 6-view (top, bottom, left, right, front, back) renderings of the original asset and emit-only renderings.

The observed defects fall into a few recurring patterns, such as emission covering the whole object or the wrong part, baked emission that is barely visible in the renders, and assets that are scenes or contain several objects.

\begin{figure*}
    \centering
       \includegraphics[width=1.0\linewidth]{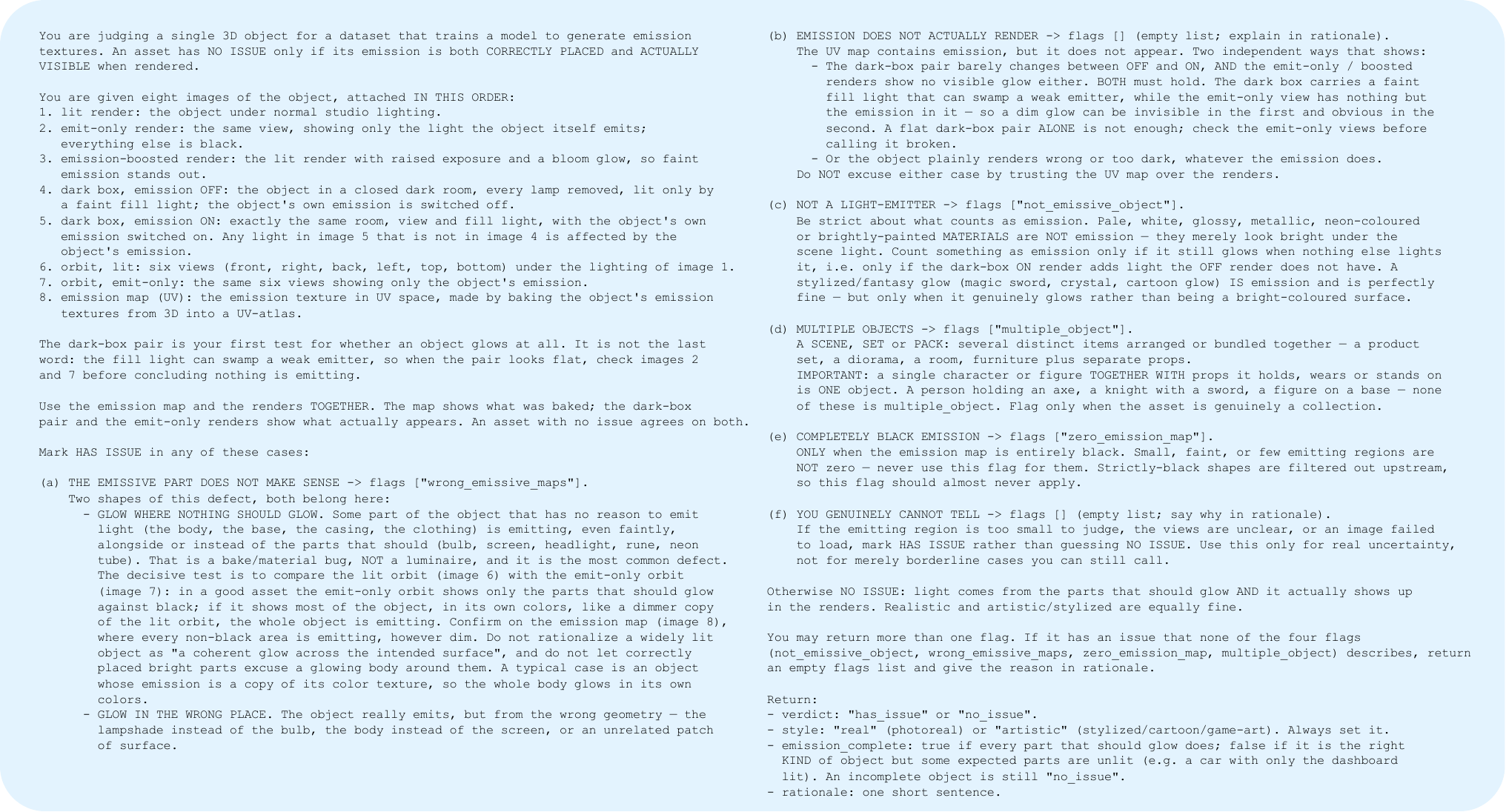}
       \caption{\textbf{VLM Prompt.} The full prompt given to GPT-5.6 Luna~\cite{openai2026gpt56} for every shape, adapted from the annotator guide. }

       \label{fig:supp_prompt}
\end{figure*}

Labeling the remaining 66k shapes by hand is not practical, so we turn the annotator guide into the prompt for GPT-5.6 Luna \cite{openai2026gpt56} (see prompt in \cref{fig:supp_prompt}).
The VLM sees the same images as the annotators. The VLM returns a decision, the defect flags, and a one-sentence rationale.

\begin{figure}
    \centering
       \includegraphics[width=1.0\linewidth]{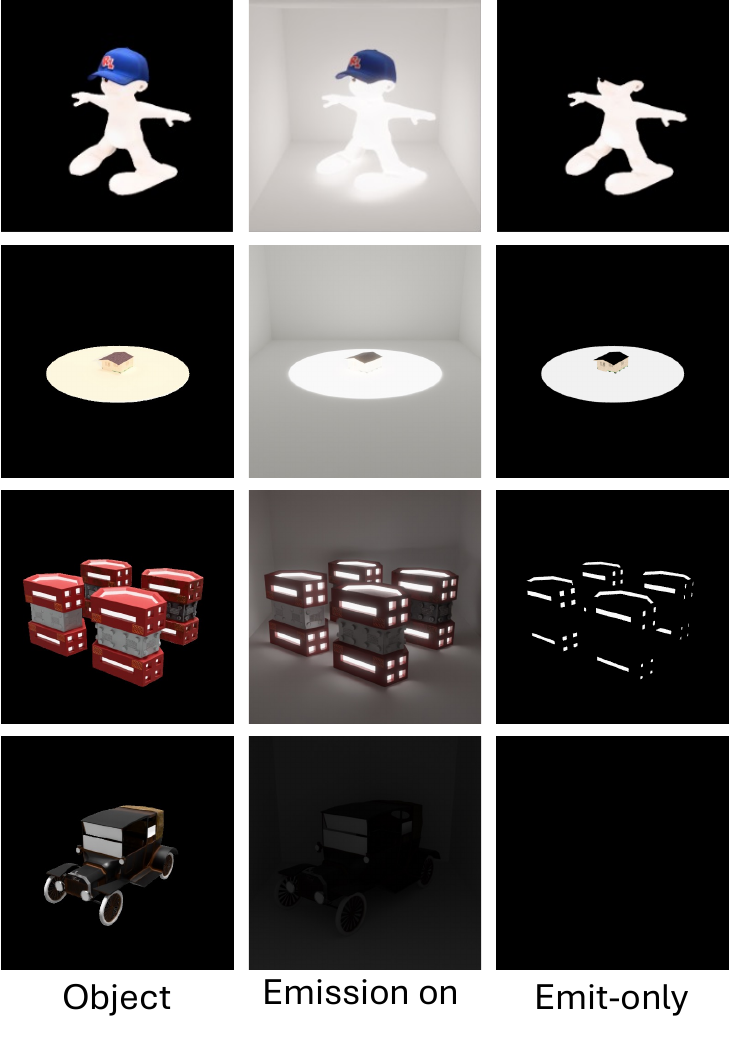}
       \caption{\textbf{Examples rejected by the VLM.} Each row from left to right shows the normal render (Object), the dark-box render with emission on, and the emit-only render.}
       \label{fig:supp_vlm_filter}
\end{figure}

To tune and test the VLM, we label another 1{,}000 random shapes; 802 of them remain after the previous stages, and we split them into 398 for tuning and 404 for testing.
On the tuning set, we revise the prompt; the main revision asks the VLM to compare the lit and emit-only renderings, since an asset whose whole body glows is a defect even when its lamps are correctly placed.
On the held-out test set, the final prompt agrees with the annotators on 86.9\% of the shapes and catches 82.8\% of the defective ones.
Applied to the 66k shapes, the VLM keeps 37{,}058 shapes.
36{,}826 of them have all representations successfully produced and form the final dataset (see~\cref{fig:supp_vlm_filter} for examples of VLM filtered data).

\subsection{Dataset Representations}
\label{sec:supp_representations}
We store every filtered asset in three representations: UV-atlas, o-voxels, and multi-view renderings.
Each representation contains the geometry, the non-emissive PBR material, and the emissive material.
The three representations are aligned by construction, since they all derive from the same normalized source asset.

\paragraph{UV-atlas.}
We re-parametrize each mesh with xatlas\footnote{\url{https://github.com/jpcy/xatlas}} into a single $512^2$ atlas.
We then bake the position, normal, occupancy mask, base color, metallic, roughness, transparency, and emission color with nvdiffrast~\cite{Laine2020diffrast}.
\paragraph{o-voxels.}
The same mesh is voxelized at a resolution of $256^3$, storing only occupied voxels. 
Each voxel carries the base color, metallic, roughness, and transparency. The emission color is stored as a separate channel set on the same voxel coordinates. The o-voxels are voxelized with the o-voxel library from TRELLIS.2~\cite{xiang2025trellis2}.
\paragraph{Multi-view renderings.}
Each asset is rendered using nvdiffrast~\cite{Laine2020diffrast} from the six orthographic cameras of Hunyuan3D-2.1~\cite{hunyuan3d2025} (front, right, back, left, top, bottom) at $512^2$.
Every view stores the position, normal, albedo, metallic--roughness, transparency, and emission maps.

\subsection{Dataset Statistics}
\label{sec:supp_statistics}
\begin{figure*}
  \centering
  \includegraphics[width=\textwidth]{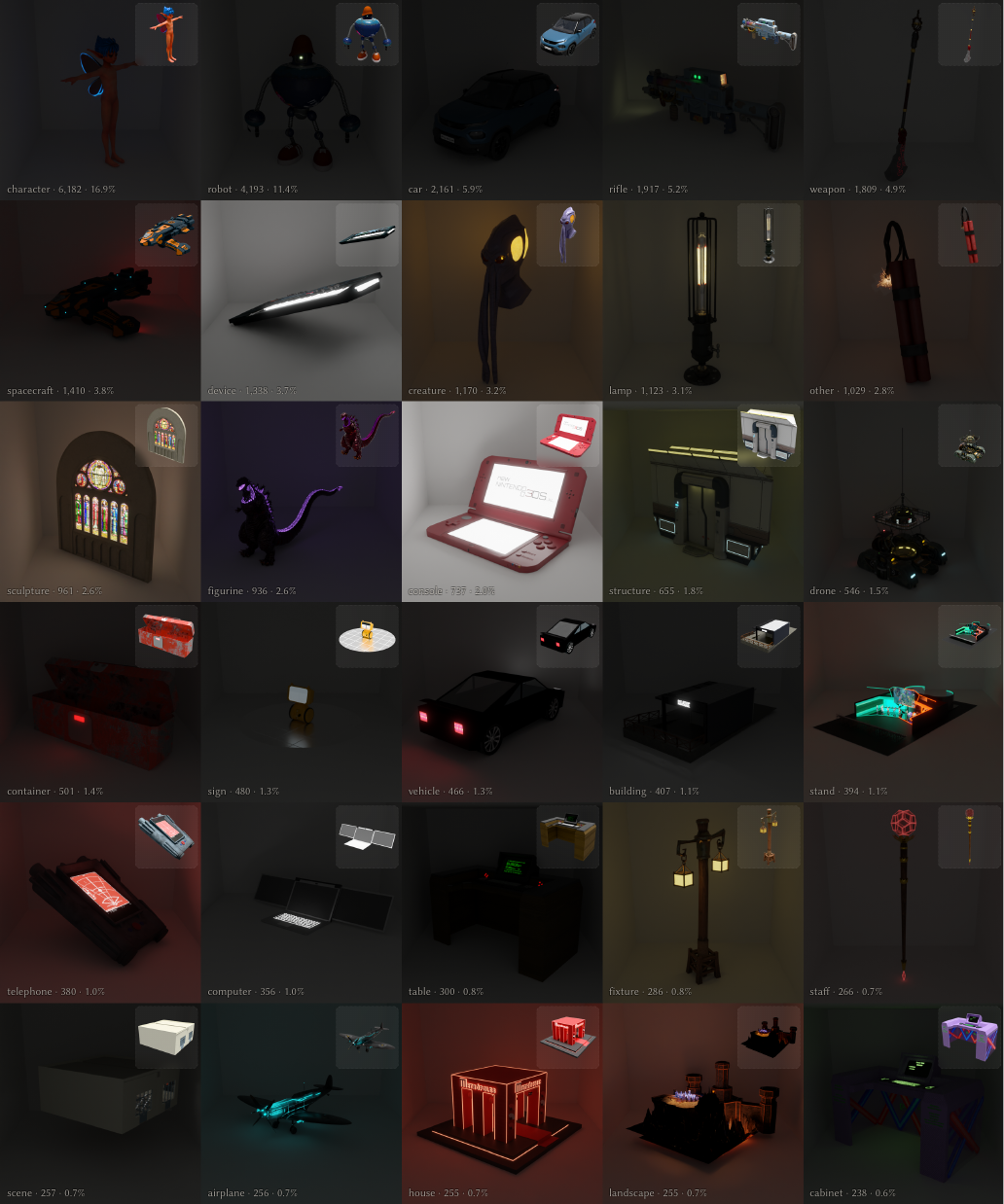}
  \caption{\textbf{Samples from the curated dataset.} Thirty assets, one drawn
  at random from each of the thirty largest categories. Each cell shows the
  asset in our dark room setting, lit by its own emission; the card in the top
  right shows the same asset under standard lighting, and the line at the
  bottom left gives the category, its shape count, and its percentage of the dataset.}
  \label{fig:supp_gallery}
\end{figure*}

In this section, we report further statistics of the dataset: the category composition, the emission strength of the source assets, and the per-shape emissive ratio.
\Cref{fig:supp_gallery} shows one random asset from each of the thirty largest categories.

\paragraph{Categories.}
\begin{figure}
    \centering
       \includegraphics[width=1.0\linewidth]{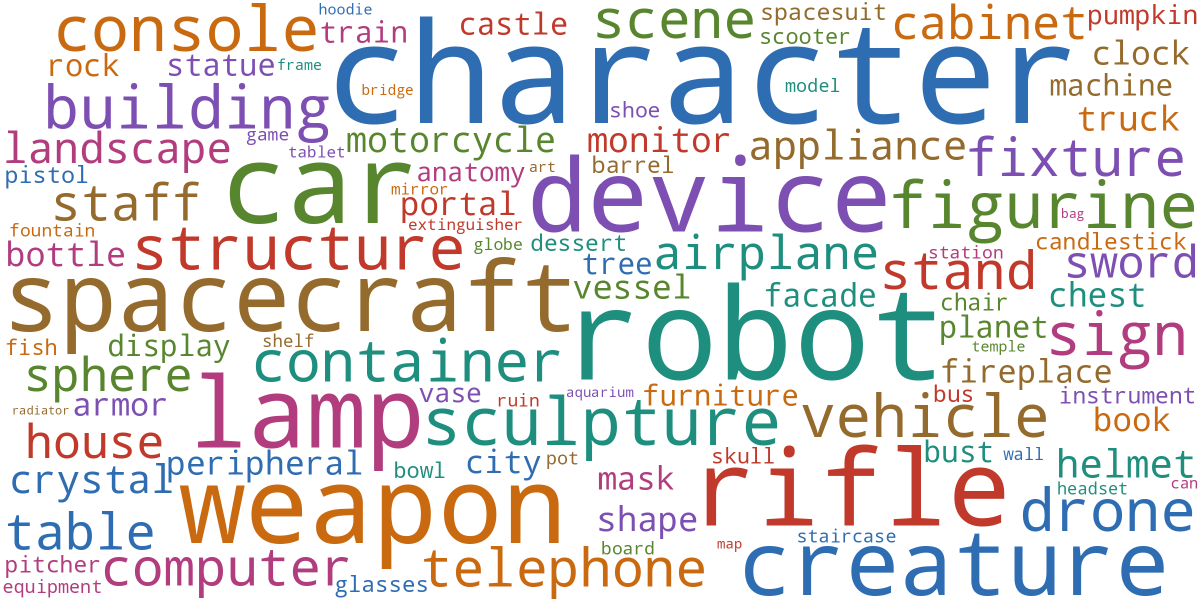}
       \caption{\textbf{Word cloud for asset categories.} The 105 categories of the final dataset. Larger words mark categories with more shapes. The \emph{other} class (3.0\%) is omitted.}
       \label{fig:supp_wordcloud}
\end{figure}

\begin{figure*}[p]
  \centering
  \includegraphics[width=\textwidth]{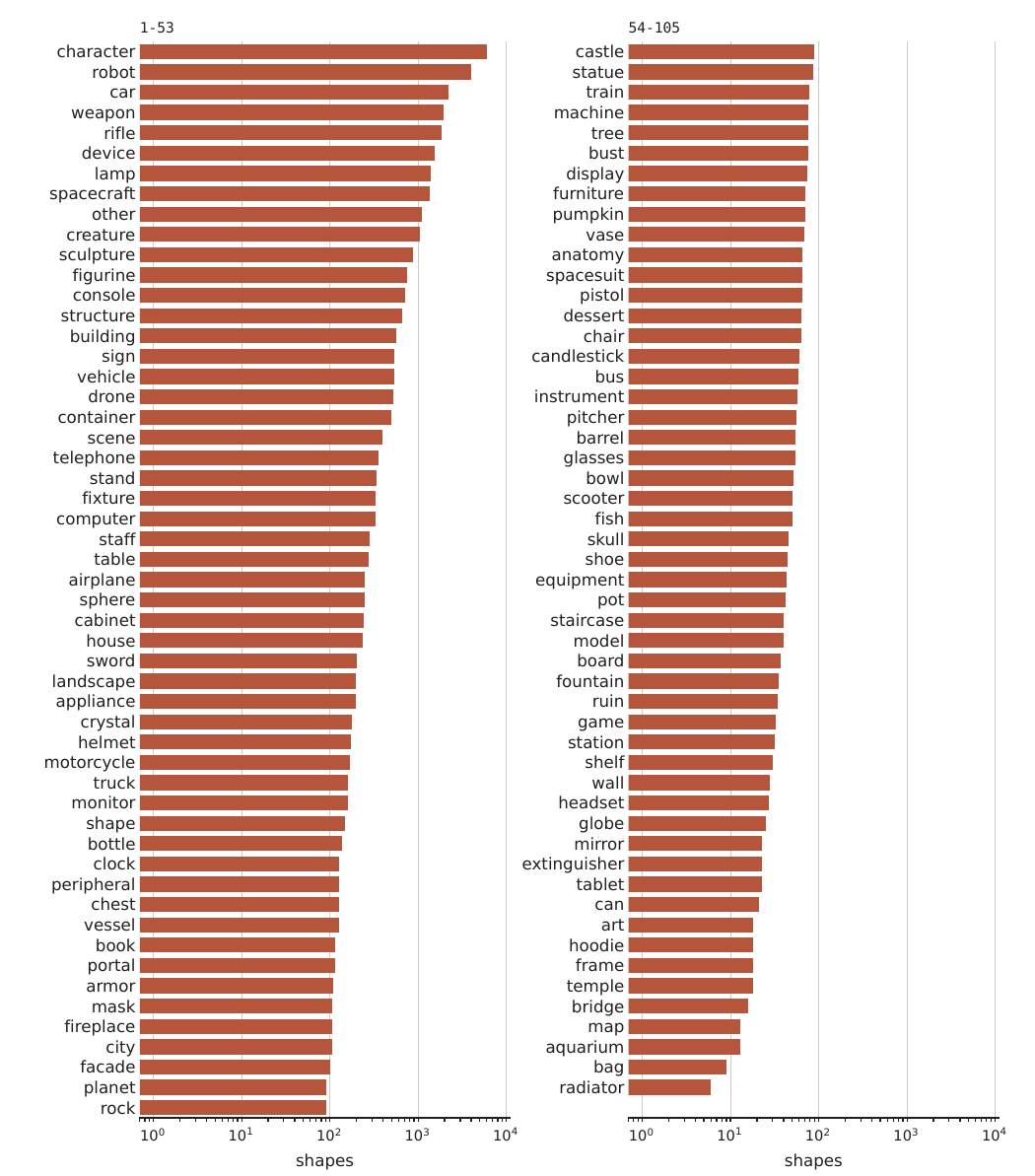}
  \caption{\textbf{Category distribution.} All 105 categories, sorted by
  size, on a log scale, split over two columns (the largest 53 on the left,
  the remaining 52 on the right) sharing one axis.}
  \label{fig:supp_categories}
\end{figure*}

To understand the composition of the dataset, we assign a category to every shape.
We use GPT-4o mini~\cite{hurst2024gpt} to describe every shape with a free-form noun phrase, given the TexVerse thumbnail and three axis-aligned (front, right, top) multi-view renderings.
We embed the descriptions with text-embedding-3-small~\cite{openaiapitextembedding3small} and cluster them with HDBSCAN~\cite{mcinnes2017hdbscan} into 552 fine-grained clusters.
We then use GPT-4o mini to name each cluster from images of five and the descriptions of thirty random members, and to merge the cluster names into broad categories (e.g., \emph{fantasy sword} and \emph{fantasy weapon} collapse into \emph{weapon}).
Every clustered shape inherits its cluster's merged name.
The shapes that fall into no cluster are classified against the merged vocabulary by GPT-4o mini, whose choices include an explicit \emph{other} class.
Our dataset spans 105 categories, with 3.0\% in \emph{other}. \Cref{fig:supp_wordcloud} shows a word cloud. The distribution is long-tailed (\cref{fig:supp_categories}): the five largest categories (character, robot, car, weapon, and rifle) hold 43.2\% of the dataset, 55 categories hold a hundred shapes or fewer, and the smallest holds six.

\paragraph{Emission strength.}
Emission strength is a scalar multiplier on the emission map.
During metadata filtering, we also read the emission strength. 
Of the 83k candidates, $\sim$80\% keep the default value of 1.
For the assets with a non-default emission strength, nearly half
use exactly 10 (see~\cref{tab:supp_emission_strength}). 
The emission strength is thus not informative for most assets, so we do not model it for our baselines.
\begin{table}
\centering
\caption{\textbf{Emission strength in the source assets.}
Per-asset maximum of the glTF emission strength over the 83k candidates. Most of them ($\sim$80\%) only have the default value of 1. $\sim$9.5\% have a value of 10. The values for the remaining 10\% are between 1 and 10.}
\label{tab:supp_emission_strength}
\small
\begin{tabular}{@{}lrr@{}}
\toprule
Emission strength & Assets & Share \\
\midrule
1 (default value)                     & 66{,}693 & 80.2\% \\
10              &  7{,}872 &  9.5\% \\
other values           &  8{,}559 &  10.3\% \\
\midrule
total                           & 83{,}124 & 100\%  \\
\bottomrule
\end{tabular}
\end{table}

\paragraph{Emissive ratio.}
\begin{figure}[t]
  \centering
  \includegraphics[width=\columnwidth]{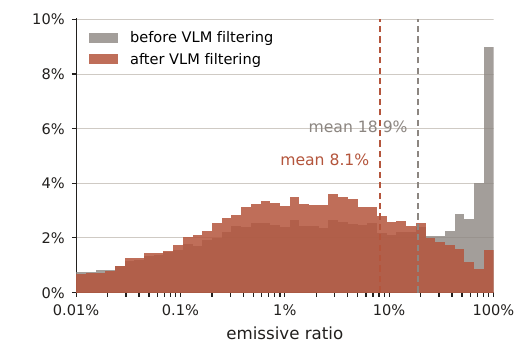}
  \caption{\textbf{Emissive ratio before and after VLM filtering.} Bars show
  the percentage of shapes within each set; the spike at 100\% in the
  pre-VLM distribution is the fully emissive shapes, most of which the VLM stage
  removes.}
  \label{fig:supp_emissive_ratio}
\end{figure}

We measure each shape's emissive ratio: the percentage of its UV-occupied texels with non-zero emission.
\Cref{fig:supp_emissive_ratio} shows its distribution before and after VLM filtering: the mean falls from 18.9\% before the VLM filtering to 8.1\% in the final dataset, since the VLM removes 87\% of the shapes whose emission covers at least half of their surface.
Emission in the dataset is sparse: the mean ratio is 8.1\% and the median 1.4\%, with the 25th and 75th percentiles at 0.3\% and 7.0\%.
Our train/val/test splits have similar mean emissive ratios (8.1\%, 8.7\%, and 7.4\%).

\section{Method and Training Details}
In this section, we describe the implementation and training details for each baseline.
\label{sec:supp_method_training_details}
\subsection{TEXGen-Emission}
\label{sec:supp_texgen}
TEXGen-Emission is our UV-domain baseline: the original TEXGen's PointUVNet~\cite{yu2024texgen} is repurposed from albedo completion to emission generation. 
The model is trained from scratch on our dataset.
Below we list the architecture changes, the training procedure, and the inference details.
\paragraph{Architecture.}
We keep TEXGen's PointUVNet~\cite{yu2024texgen} and change only its input channels and the resolution.
The original TEXGen takes a 10-channel UV input: the noised albedo, the position map, the partial albedo, and the occupancy mask.
We replace it with our 13-channel stack
$\{z_t \Vert P \Vert a \Vert m \Vert r \Vert \alpha \Vert M\}$.
Accordingly, we resize the two layers that read it, the input convolution and the adaptive-skip
convolution.
The output remains the 3-channel velocity of $z_t$.
The image condition is the CLIP embedding of the asset's thumbnail ($224^2$), and the text branch receives the CLIP embedding of the fixed prompt ``emission generation''.
To save compute, we run the model at a $256^2$ UV resolution instead of the
original $1024^2$, and shorten the serialized-attention patch lengths from
256/256/512/1024 to 128/128/256/256 because of the resolution change.
\paragraph{Training.}
TEXGen releases inference code only, so we implement the training ourselves.
We do not initialize from the released TEXGen weights, since the task shifts
from albedo completion to emission generation.
The inputs are the $512^2$ baked atlases downsampled to $256^2$ and
normalized to $[-1,1]$.
We follow TEXGen's flow-matching formulation.
The noised emission is $z_t = (1-t)\,\epsilon + t\,z_0$ with
$\epsilon \sim \mathcal{N}(0, I)$, and the network predicts the velocity
$v = z_0 - \epsilon$.
Timesteps are sampled as $t = u^2$ with $u \sim \mathcal{U}(0,1)$.
The loss is $\ell_2 + 0.5\,\ell_1$ between the predicted and target velocity within the occupied texels.
We drop TEXGen's differentiable-rendering loss and supervise in UV space
only.
For classifier-free guidance, the image and text embeddings are zeroed with
probability $0.1$.
We use AdamW ($\text{lr}=5\times10^{-4}$, weight decay $0.01$) with a cosine schedule, gradient clipping at $1.0$, \texttt{bf16} mixed precision, and a global batch size of 128, and keep an EMA of the weights (decay $0.9999$) that is used at inference.
The model is trained for 35{,}625 steps on four H100 GPUs.

\paragraph{Inference.}
We integrate the velocity with 50 Euler steps at a classifier-free guidance
scale of 2.0, and undo the $[-1,1]$ normalization on the occupied texels.

\subsection{TRELLIS.2-Emission}
\label{sec:supp_trellis2}
TRELLIS.2-Emission is our voxel-domain baseline: the PBR generation stage of TRELLIS.2~\cite{xiang2025trellis2} is repurposed for emission generation.
Both modules are fine-tuned from the released PBR weights on our dataset.
Below we detail the architecture, training,
and inference.
\paragraph{Architecture.}
\label{sec:TRELLIS2_arch}

\begin{table}
\centering
\caption{\textbf{Emission VAE ablation.}
 Reconstruction quality of the ground-truth emission with the pretrained PBR VAE and our emission VAE on the test set. MAE is measured on the ground-truth emissive voxels; IoU is computed over all occupied voxels.}
\label{tab:supp_vae_ablation}
\small
\begin{tabular}{lcc}
\toprule
VAE & MAE $\downarrow$ & IoU $\uparrow$ \\
\midrule
PBR VAE & 0.0461 & 0.694 \\
\textbf{Emission VAE} & \textbf{0.0330} & \textbf{0.746} \\
\bottomrule
\end{tabular}
\end{table}

Our emission VAE and emission DiT follow the architecture of TRELLIS.2's PBR VAE and PBR DiT.
We modify only the inputs and outputs for emission generation.
The PBR VAE encoder reads the 6-channel non-emissive material, while ours reads the 3-channel emission $L_{\text{e}}$.
The PBR VAE decoder reads its own 32-channel latent, while ours additionally reads the 32-channel latent $\text{z}_{\text{PBR}}$ from the PBR VAE encoder.
Our emission VAE outputs 3-channel emission instead of the 6-channel material.
We use a dedicated emission VAE because it empirically outperforms the original PBR VAE (\cref{tab:supp_vae_ablation}).
The PBR DiT's input is $[\,\text{x}_t \,\Vert\, \text{z}_{\text{shape}}\,]$ (64 channels).
Our emission DiT's input is $[\,\text{x}_t \,\Vert\, \text{z}_{\text{shape}} \,\Vert\, \text{z}_{\text{PBR}}\,]$ (96 channels), and it predicts the velocity of the emission latent
$\text{z}_{\text{e}}$ instead of the PBR latent $\text{z}_{\text{PBR}}$.
$\text{z}_{\text{shape}}$ and $\text{z}_{\text{PBR}}$ come from the frozen
pretrained encoders.
Four layers change shape: the VAE input and output projections, the decoder latent projection, and the DiT input projection.
\paragraph{Training.}
We fine-tune both the emission VAE and emission DiT from the pretrained PBR VAE and PBR DiT.
For the four layers above that change shape, the emission VAE's input and output projections take the pretrained weights of the albedo channels, and the decoder's latent projection keeps the pretrained weights for $\text{z}_{\text{e}}$ and zero-initializes the new weights for $\text{z}_{\text{PBR}}$.
The DiT's 96-channel input weight is randomly initialized.
We train the emission VAE first.
The inputs are the $256^3$ o-voxels, with the emission and material attributes normalized from 8-bit values to $[-1,1]$.
The loss is $\ell_1$ between the reconstructed and target emission over the occupied voxels, plus a KL term with weight $10^{-7}$.
We use AdamW ($\text{lr}=10^{-4}$, no weight decay) with a cosine schedule, gradient clipping at the 95th percentile of recent gradient norms (at most $1.0$), \texttt{fp16} mixed precision, and a global batch size of 64.
The VAE is trained for 35{,}800 steps on four H100 GPUs, and we keep the checkpoint with the lowest validation loss.
We then encode the emission of every shape with the emission VAE.
We train the DiT with the encoded emission latent, the PBR latent from the pretrained PBR VAE, and the shape latent from the pretrained shape VAE.
We use TRELLIS.2's flow-matching objective: the noised latent is
$\text{x}_t = (1-t)\,\text{x}_0 + (\sigma_{\min} + (1-\sigma_{\min})\,t)\,\epsilon$
with $\sigma_{\min}=10^{-5}$ and $t \sim \mathcal{U}(0,1)$.
The loss is $\ell_2$ on the velocity target $(1-\sigma_{\min})\,\epsilon - \text{x}_0$.
For classifier-free guidance, the image tokens are zeroed with probability $0.1$; $\text{z}_{\text{shape}}$ and $\text{z}_{\text{PBR}}$ are never dropped.
We use AdamW ($\text{lr}=10^{-4}$, $\beta=(0.9, 0.95)$, weight decay $0.01$) with a 1{,}000-step linear warm-up and a constant learning rate thereafter, the same gradient clipping, \texttt{bf16} mixed precision, and a global batch size of 32.
The DiT is trained for 50{,}000 steps on four H100 GPUs.
Both stages keep an EMA of the weights (decay $0.9999$), but we evaluate with the online weights.
\paragraph{Inference.}
Given a shape, we encode its geometry and non-emissive material with the frozen pretrained shape and PBR VAE encoders to obtain $\text{z}_{\text{shape}}$ and $\text{z}_{\text{PBR}}$. We extract the DINOv3~\cite{simeoni2025dinov3} features of its thumbnail as the image condition.
We integrate the velocity from Gaussian noise with 50 Euler steps at a classifier-free guidance scale of 3.0, decode $[\,\text{z}_{\text{e}} \,\Vert\, \text{z}_{\text{PBR}}\,]$ with the emission VAE decoder, and undo the $[-1,1]$ normalization on the occupied voxels.
\subsection{Hunyuan3D-Emission}
 \label{sec:supp_hunyuan}
Hunyuan3D-Emission is our multi-view baseline: the multi-view diffusion model of Hunyuan3D-2.1 Paint~\cite{hunyuan3d2025} is repurposed from albedo and metallic--roughness generation to emission generation.
The model is fine-tuned from the released weights.
\paragraph{Architecture.}
We keep Hunyuan3D-2.1 Paint's multi-view UNet and change only its generation branches and input channels.
The original model generates albedo and metallic--roughness as two branches that share the UNet, each with its own learned prompt token and attention projections.
We keep a single emission branch and initialize its token and projections from the albedo branch.
The original per-view input is a 12-channel latent stack: the noised latent, the normal latent, and the position latent.
We replace it with the 24-channel stack $\{z_t \Vert n \Vert P \Vert a \Vert (m,r) \Vert \alpha\}$.
Accordingly, we widen the input convolution from 12 to 24 channels and zero-initialize the new weights.
The input reference image is the asset's thumbnail.
\paragraph{Training.}
We fine-tune from the released Hunyuan3D-2.1 Paint weights.
The inputs are the $512^2$ multi-view renderings normalized to $[-1,1]$.
We follow Hunyuan3D-2.1 Paint's DDPM formulation with $v$-prediction~\cite{salimans2022progressive}.
The loss is $\ell_2$ between the predicted and target velocity over the six views.
We use a single reference image and drop the dual-reference consistency loss.
We keep the released recipe for weight freezing and condition dropout (probability $0.1$).
We use AdamW ($\text{lr}=7\times10^{-5}$, weight decay $0.01$) with the released warm-restart cosine schedule, gradient clipping at $1.0$, \texttt{bf16} mixed precision, a global batch size of 16, and no EMA.
We train for 50{,}000 steps on four H100 GPUs. 
We select the best checkpoint based on the validation split.
\paragraph{Inference.}
We sample the six emission views with 30 Euler-ancestral steps without classifier-free guidance, and decode them with the Stable Diffusion VAE.
We back-project the views onto the $512^2$ UV atlas following Hunyuan3D-2.1 Paint's cosine-weighted blending of the visible views, but without its inpainting step.
Unobserved texels get zero emission. Inpainting the unobserved texels may spread emission into regions that should not glow.
Since the views are rendered on a gray background, we subtract the gray from silhouette pixels before projection.
\subsection{SegviGen-Emission}
\label{sec:supp_segvigen}
SegviGen-Emission is our segmentation baseline: SegviGen~\cite{segvigen}, a part-segmentation model built on the TRELLIS.2 PBR DiT, is repurposed from part segmentation to emissive-region segmentation.
The model is fine-tuned from the released full-segmentation weights.
\paragraph{Architecture.}
We keep SegviGen's DiT, which is the TRELLIS.2 PBR DiT of \cref{sec:TRELLIS2_arch}, and change only the target.
The original SegviGen generates a per-voxel part coloring written into the albedo slot of the 6-channel material.
We replace it with the binary emissive mask $M$, written as $[\,M, M, M, 0, 1, 1\,]$; a voxel is emissive if any channel of its ground-truth emission is non-zero.
The target is encoded into the latent $\text{z}_{M}$ with the pretrained PBR VAE encoder and decoded with the pretrained PBR VAE decoder, both frozen.
The inputs and conditions are unchanged from SegviGen: $\text{z}_{\text{shape}}$, $\text{z}_{\text{PBR}}$, and the DINOv3~\cite{simeoni2025dinov3} features of the asset's thumbnail rendering.
\paragraph{Training.}
We fine-tune from SegviGen's full-segmentation checkpoint.
The inputs are the $256^3$ o-voxels, with the mask and material attributes normalized to $[-1,1]$ before encoding.
We use the same flow-matching objective as TRELLIS.2-Emission, with the $\ell_2$ loss on the velocity of $\text{z}_{M}$.
We do not use condition dropout, so there is no classifier-free guidance.
We use AdamW ($\text{lr}=10^{-5}$, weight decay $0.01$) with a constant learning rate, the same gradient clipping as TRELLIS.2-Emission, \texttt{bf16} mixed precision, and a global batch size of 48, and keep an EMA of the weights (decay $0.999$) that is not used at inference.
The model is trained for 44{,}722 steps on four H100 GPUs.
We select the best checkpoint based on the validation split.
\paragraph{Inference.}
Given a shape, we obtain $\text{z}_{\text{shape}}$, $\text{z}_{\text{PBR}}$, and the DINOv3 tokens of its thumbnail as in TRELLIS.2-Emission.
We integrate the velocity from Gaussian noise with 12 Euler steps without classifier-free guidance, decode $\text{z}_{M}$ with the PBR VAE decoder, and undo the $[-1,1]$ normalization on the albedo channels.
We threshold the result at $1/255$ to obtain the predicted mask $M$, and take the emission as $M \odot a$ with the input albedo $a$.

\section{Evaluation Protocol Details}
\label{sec:supp_evaluation_protocol_details}
This section provides details on the sampled point cloud evaluation, the process for dark room rendering, and the setup of the user study.

\begin{figure}
    \centering
       \includegraphics[width=1.0\linewidth]{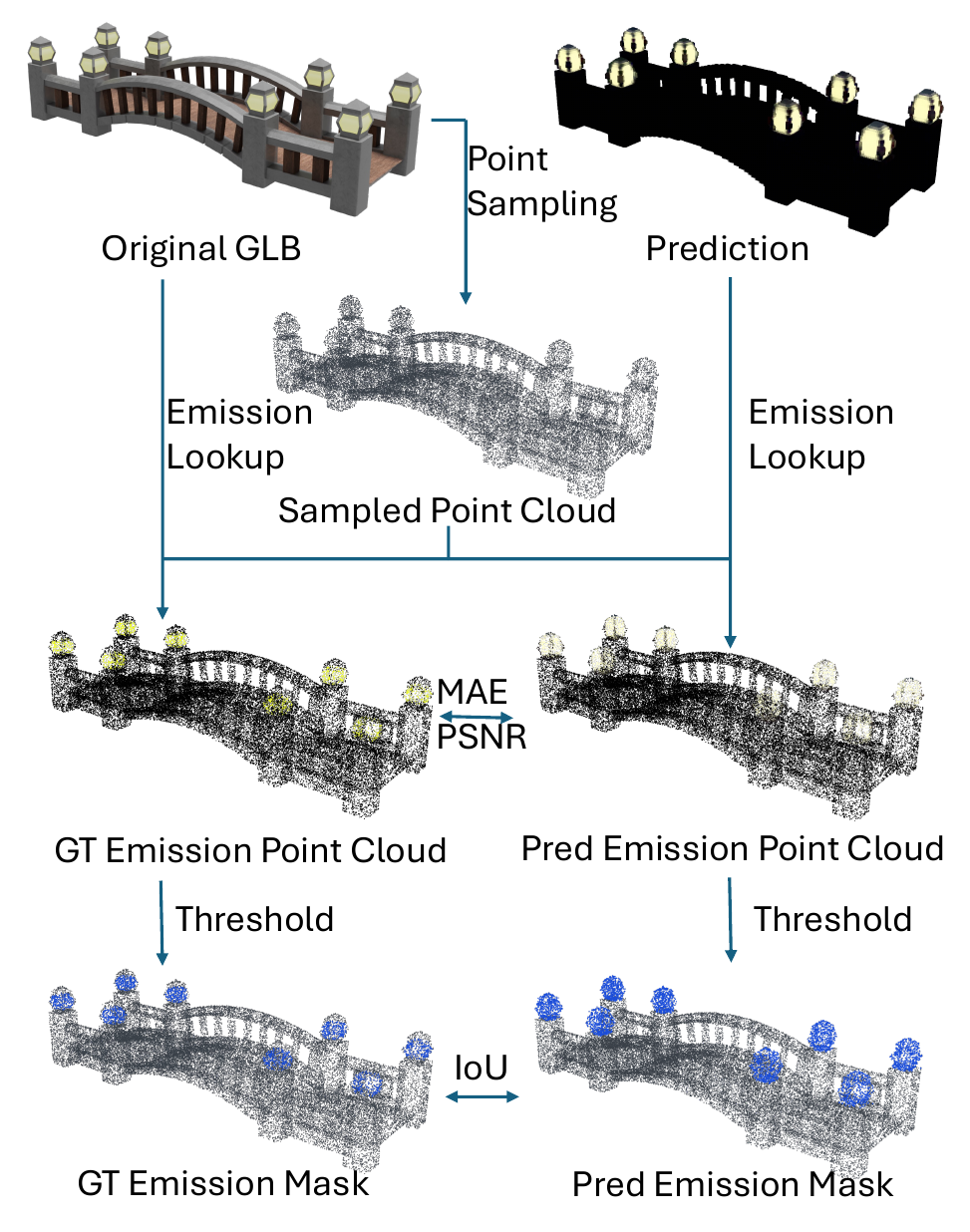}
       \caption{\textbf{Sampled point cloud evaluation.} We first sample 50{,}000 points from the original 3D asset. 
       For the ground-truth emission, we read the color directly at the
       corresponding location on the original asset.
       For the predicted emission, we use nearest neighbor lookup. 
       With the ground truth and the predicted emission, we apply a threshold to obtain the emission mask. MAE and PSNR compare the emission point clouds, and IoU compares the thresholded masks.}
       \label{fig:point_eval}
\end{figure}

\paragraph{Sampled Point Cloud Evaluation.}
\label{sec:supp_point_eval}
For each asset, we sample 50{,}000 points on the source mesh, uniformly by
surface area, with a fixed random seed.
All methods and all their random seeds are evaluated on the same points
(\cref{fig:point_eval}).
The points are sampled from the normalized asset of \cref{sec:supp_dataset},
in which the three representations are baked, so no further alignment is
needed.
Each point is then assigned a reference emission and a predicted emission.
The reference is the source asset's emission color at that point.
The prediction is read from each method's output by a nearest-neighbor lookup.
The lookup is the nearest occupied voxel for TRELLIS.2-Emission and SegviGen-Emission, and the nearest atlas texel for TEXGen-Emission and Hunyuan3D-Emission, whose six generated views are first back-projected into a UV atlas.

\paragraph{Dark Room Rendering.}
All results are rendered on the source asset with Blender.
The predictions are transferred to the asset's own UV textures and used as the
emission color of its materials.
The asset is placed in a box with gray walls.
The asset is lit only by its own emission and a
weak white world light.
The world light keeps the object slightly visible, so that we can tell what
the object is when the emission is off.
Each asset has one fixed camera, shared by the ground truth and all baselines.
For the user study videos, the emission strength ramps from 0 to 8 over 24
frames at 8\,fps.
The qualitative results in the main paper are rendered at emission strength 4.2.

\begin{figure}
    \centering
       \includegraphics[width=1.0\linewidth]{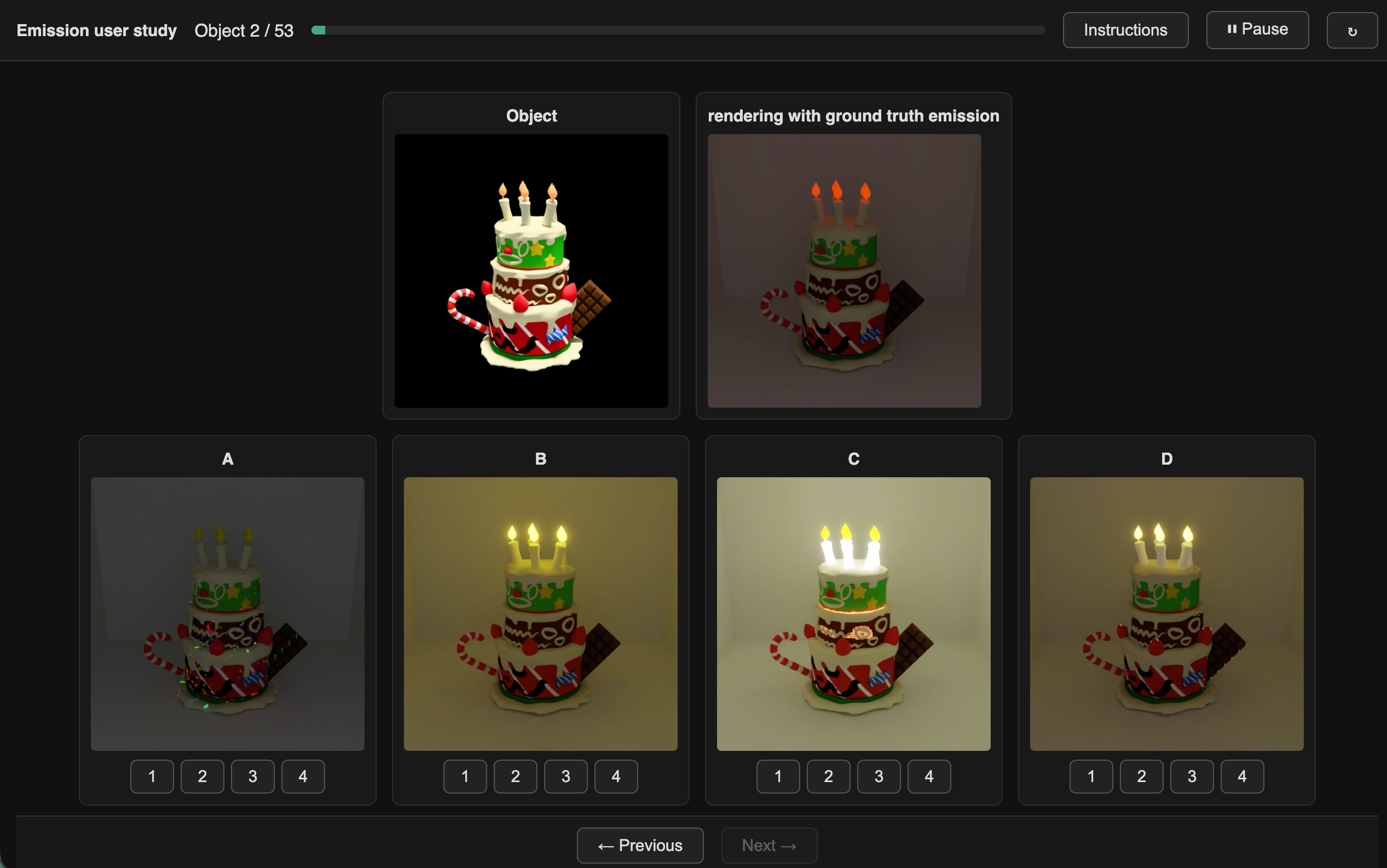}
        \caption{\textbf{User study interface.} The top row is the reference: the object under standard lighting and the video with its ground-truth emission.
         The bottom row shows the four methods in a random order, labeled A--D.
          Participants rank them from 1 to 4 based on the plausibility of the emission. The participants can refer to the references as a guide.}
         \label{fig:user_study_snapshot}
\end{figure}

\paragraph{User Study.}
\Cref{fig:user_study_snapshot} shows the interface participants use.
We have 14 participants, and each of them ranks all four baselines on the same 53 assets.
This results in 742 rankings in total.
The mean rank and the top-1 rate reported in the main paper are computed over all rankings.
The mean rank is the average rank of a baseline.
The top-1 rate is the fraction of rankings in which the baseline is ranked first.

\section{Additional Results}
\label{sec:supp_additional_results}

\begin{table}
\centering
\caption{\textbf{Emissive-mask precision, recall, and F1.} We report the mean over five random seeds of the generator. The emissive masks are obtained by a $1/255$ threshold. TRELLIS.2-Emission is best on all four mask-based metrics, but no method localizes the emissive region well: even the best F1 is only 0.415.}
\label{tab:supp_prf}
\begin{tabular}{@{}lcccc@{}}
\toprule
Method & IoU $\uparrow$ & Prec. $\uparrow$ & Rec. $\uparrow$ & F1 $\uparrow$ \\
\midrule
TEXGen-Emission    & 0.092 & 0.135 & 0.192 & 0.159 \\
TRELLIS.2-Emission & \textbf{0.271} & \textbf{0.361} & \textbf{0.487} & \textbf{0.415} \\
Hunyuan3D-Emission & 0.117 & 0.153 & 0.411 & 0.223 \\
SegviGen-Emission  & 0.147 & 0.200 & 0.342 & 0.252 \\
\bottomrule
\end{tabular}
\end{table}

\begin{table}
\centering
\caption{\textbf{MAE and PSNR on ground-truth emissive points (Mask) and on all points (All).} We report the mean over five random seeds of the generator. A point is emissive if its ground-truth emission exceeds $1/255$. TRELLIS.2-Emission is slightly better, but no method matches the artist-authored emission well.}
\label{tab:supp_mask_magnitude}
\scalebox{0.9}{
\begin{tabular}{@{}lcccc@{}}
\toprule
& \multicolumn{2}{c}{Mask} & \multicolumn{2}{c}{All} \\
\cmidrule(lr){2-3} \cmidrule(l){4-5}
Method & MAE $\downarrow$ & PSNR $\uparrow$ & MAE $\downarrow$ & PSNR $\uparrow$ \\
\midrule
TEXGen-Emission    & 0.318 & 9.62 & 0.025 & 25.10 \\
TRELLIS.2-Emission & \textbf{0.264} & \textbf{11.07} & \textbf{0.021} & \textbf{26.81} \\
Hunyuan3D-Emission & 0.312 & 10.00 & 0.022 & 26.46 \\
SegviGen-Emission  & 0.299 & 10.39 & 0.041 & 24.50 \\
\bottomrule
\end{tabular}
}
\end{table}

\begin{figure}
    \centering
       \includegraphics[width=0.8\linewidth]{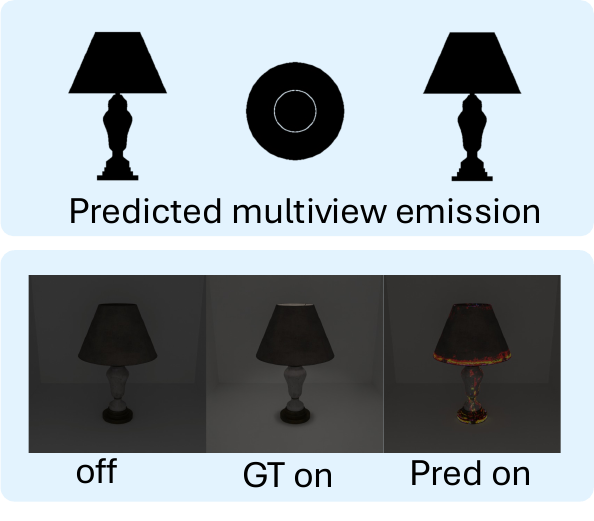}
       \caption{\textbf{Occlusion failure of the multi-view baseline.} An emissive surface hidden in the rendered views is never observed by the model, so its correct emission can not be generated. Here, the bulb of the lamp is occluded; the predicted emission (top) is nearly dark, and the lamp fails to light the dark room in the rendering when we have the light on (bottom).}
       \label{fig:multiview_failure}
\end{figure}

\begin{figure}
    \centering
       \includegraphics[width=1.0\linewidth]{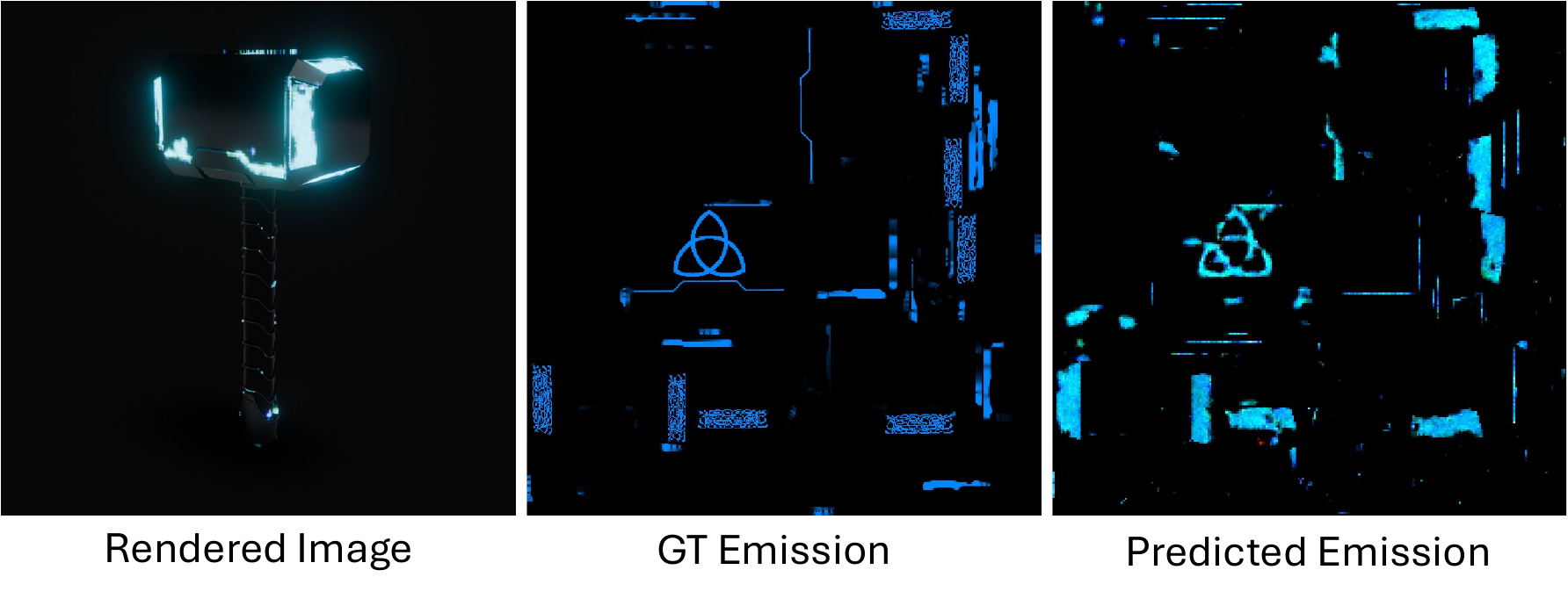}
       \caption{\textbf{Fracturing artifacts in the TEXGen-Emission model.} Emission predicted in UV space leaks into nearby UV islands, which map back to surface regions that should not emit.}
       \label{fig:texgen_fracture}
\end{figure}

\begin{figure}
  \centering
  \includegraphics[width=0.8\columnwidth]{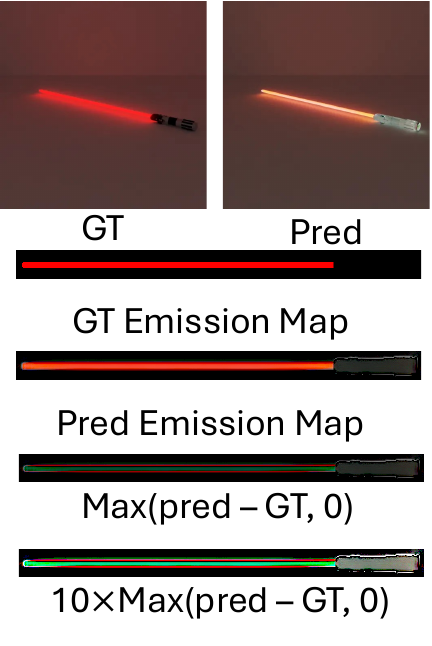}
   \caption{\textbf{Noise artifacts of Hunyuan3D-Emission.} Compared with the dark room rendering of the ground truth (1st row left), the predicted hilt of the lightsaber glows white although it should not emit (1st row right). 
   In one view of the multi-view emission output, the prediction (3rd row) differs from the ground truth (2nd row) only faintly, but the positive difference $\max(\text{pred}-\text{GT},0)$ (4th row), amplified $10\times$ (5th row), reveals non-zero emission over the whole non-emissive hilt.}
  \label{fig:supp_hy_noise}
\end{figure}

We provide additional results: precision, recall, and F1 of the predicted emissive mask (\cref{tab:supp_prf}), and MAE and PSNR restricted to the ground-truth emissive points (\cref{tab:supp_mask_magnitude}).
On the predicted mask, TRELLIS.2-Emission performs best among the baselines, leading in IoU, precision, recall, and F1.
For the emission color, no method performs well within the ground-truth mask: PSNR stays below 12\,dB for all baselines, far below their all-point values.

\begin{figure*}
  \centering
  \includegraphics[width=\textwidth]{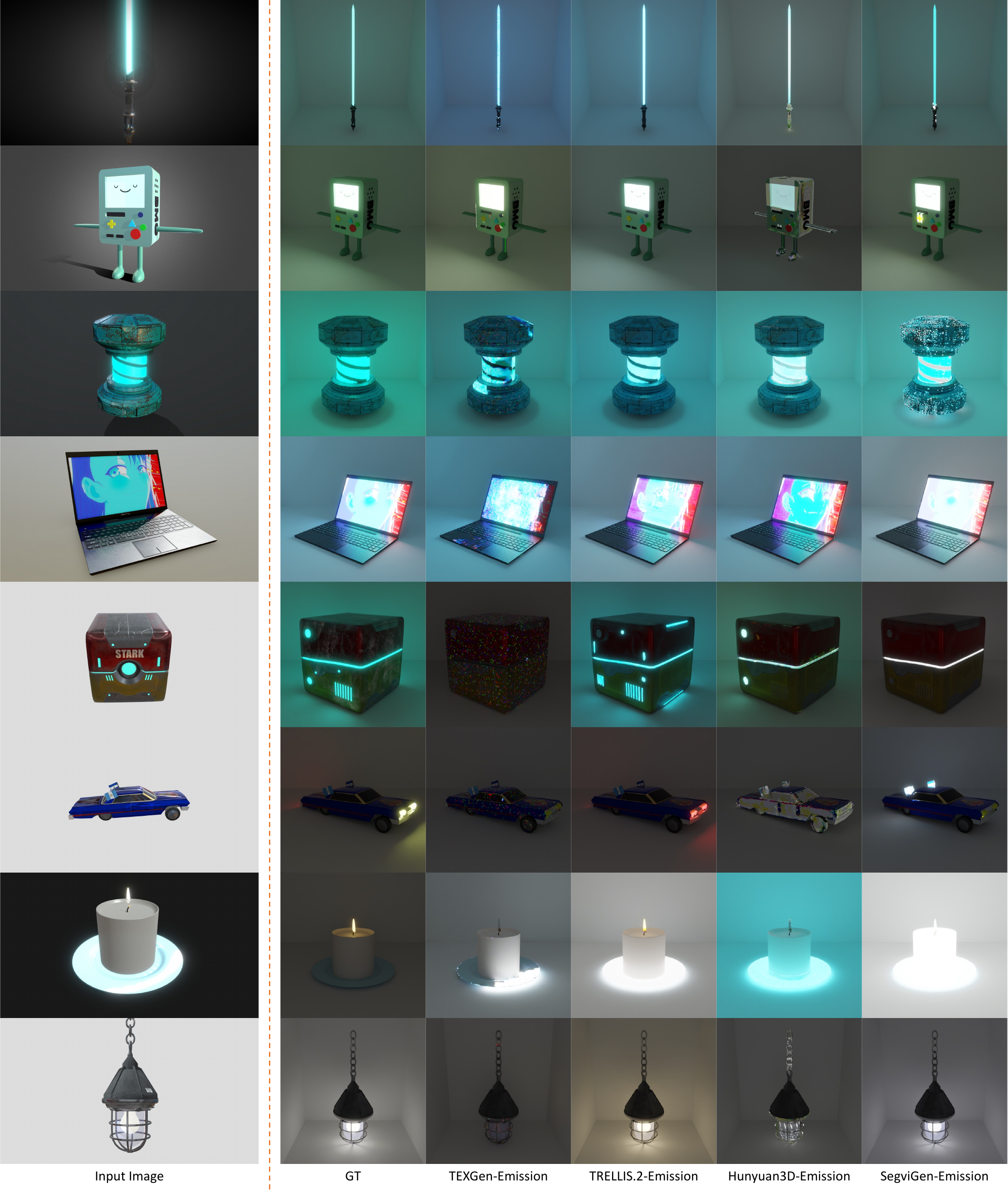}
  \caption{\textbf{Additional qualitative results.} The leftmost column is the input image, GT is the rendering with the ground-truth emission, and each object is rendered in our dark room setting (\cref{sec:supp_evaluation_protocol_details}), lit by its predicted emission.}
  \label{fig:supp_additional_qual}
\end{figure*}

\Cref{fig:supp_additional_qual} shows more qualitative results. We show these failure cases: \textbf{multi-view generation misses occluded emitters} (\cref{fig:multiview_failure}), \textbf{UV domain generation fractures across UV islands} (\cref{fig:texgen_fracture}), and the noise artifact of Hunyuan3D-Emission (\cref{fig:supp_hy_noise}).

\section{Limitations and Future Work}

\label{sec:supp_failre_cases_limitations}

Our quantitative metrics compare each prediction against a single artist-authored reference, so a plausible prediction that differs from the reference is penalized.
The only input that reveals the intended emission is the thumbnail rendering: when it shows the asset in its lit state, the task reduces to transferring the observed emission onto the surface.
Otherwise, the correct emission remains under-determined.
For such cases, a future direction is an evaluation that accounts for multiple plausible answers.

Our baselines generate emission conditioned on the full input of the benchmark: the geometry, the non-emissive material properties, and the thumbnail rendering.
Generation conditioned only on the geometry, or only on a text or image prompt, is not evaluated.
Extending the benchmark to these settings is another direction for future work.

Our dataset is sourced solely from TexVerse~\cite{zhang2025texverse}.
Extending it to other asset sources is a natural future direction.

\end{document}